%% file: main.tex
\documentclass[11pt]{article}
\usepackage[preprint]{acl}
\usepackage{times}
\usepackage{latexsym}
\usepackage[T1]{fontenc}
\usepackage[utf8]{inputenc}
\usepackage{microtype}
\makeatletter\providecommand\MT@suspend@tagging{}\makeatother % older LaTeX kernels: a newer microtype leaves this undefined in lists
\usepackage{graphicx}
\usepackage{booktabs}
\usepackage{amsmath}
\usepackage{xcolor}
\makeatletter
\newcommand\appflush{\par\if@firstcolumn\else\ifx\@deferlist\@empty\else\ifdim\dimexpr\pagegoal-\pagetotal\relax<0.4\textheight \mbox{}\newpage\fi\fi\fi}
\makeatother
\newenvironment{tightlist}{\begin{list}{\textbullet}{\setlength{\leftmargin}{1.1em}\setlength{\labelwidth}{0.7em}\setlength{\labelsep}{0.4em}\setlength{\itemsep}{1pt}\setlength{\topsep}{2pt}\setlength{\parsep}{0pt}}}{\end{list}}
\newcounter{limctr}
\input{numbers}   % generated by paper_numbers.py: every number in this manuscript is a macro
\input{tables}    % generated by paper_tables_tex.py: every table body


\def\vdEND{\vdEND}
\def\vdcheck#1#2\vdEND{\ifx#1\missingnum\missingnum\else\ifnum\pdfstrcmp{#1#2}{TRUE}=0 holds\else fails\fi\fi}
\newcommand{\vd}[1]{\expandafter\vdcheck#1\vdEND}   % verdict macro (TRUE/FALSE) -> holds/fails; a placeholder prints as is
\newcommand{\knew}{\textit{knew}}
\newcommand{\flipk}{flip$\mid$\knew}
\newcommand{\cuepos}[1]{\textcolor{red!65!black}{\textsf{[#1]}}}   % cue-position tags in the prompt figure
\newcommand{\gemma}{Gemma}\newcommand{\gemmafull}{Gemma-2-2B-it}
\newcommand{\llama}{Llama}\newcommand{\llamafull}{Llama-3.1-8B-Instruct}

\title{Consistency Training Suppresses Cue Following\\Without Teaching Cue Selectivity}

\author{
\textbf{Prakhar Gupta\textsuperscript{1,2}}\thanks{Email: \texttt{prakharg@umich.edu}.} \quad
\textbf{Terry Jingchen Zhang\textsuperscript{2,3,4}} \quad
\textbf{Zhijing Jin\textsuperscript{2,4,5}} \\[0.15em]
\textsuperscript{1}University of Michigan \quad
\textsuperscript{2}Jinesis Lab, University of Toronto \& Vector Institute\\
\textsuperscript{3}University of Oxford \quad
\textsuperscript{4}EuroSafeAI\\
\textsuperscript{5}Max Planck Institute for Intelligent Systems, T\"{u}bingen, Germany
}

\begin{document}
\maketitle
\ifnum\vIsMerged=1
\noindent\fbox{\parbox{\dimexpr\columnwidth-2\fboxsep}{\textcolor{red}{\textbf{DRAFT.} Every number below is drawn from the superseded merged-adapter pipeline (\vBasisNote). The manuscript is regenerated from the unmerged results before any use; directional claims are re-checked against CHECKLIST.md.}}}
\fi

\begin{abstract}Language models often change answers because of a cue, a hint such as ``I believe the answer is B'' that gives no valid evidence. Consistency training (CT) trains a model to answer the same with and without the cue. Four CT methods apply this idea to output tokens (BCT), residual activations (ACT), attention patterns (AttCT) and sampled-answer rates (RMCT). Prior work suggests some learn different solutions. We train all four on \gemmafull{} and \llamafull{} with the same data. We test predictions written in advance and inspect the models with a known circuit that carries cues to the answer. All four make the trained cue less effective, and their fixes share three properties. First, correct cues lose about as much influence as wrong ones. Second, the fix is usually large for cues before the answer options, as in training, and much smaller after them. \llama{} RMCT depends least on position. Third, the circuit heads still carry most remaining cue-induced errors. Making BCT targets depend on the cue's correctness mostly weakens the fix. CT thus weakens the trained cue without teaching models when a cue is right. Evaluations with only wrong cues at the trained position miss the first two properties.
\end{abstract}

\section{Introduction}

\begin{figure*}[t]
\centering
\includegraphics[width=\textwidth]{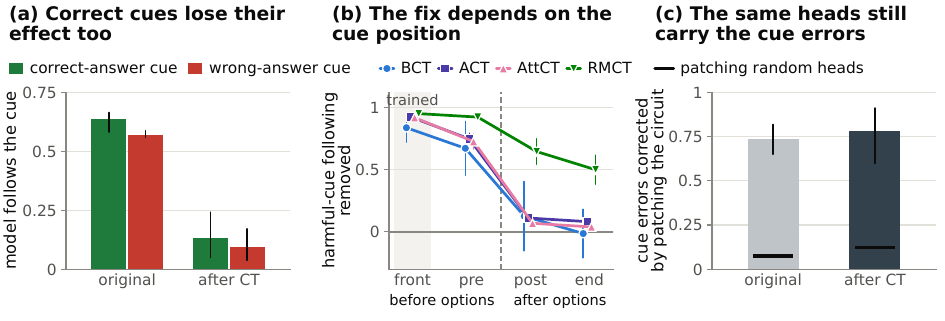}
\caption{\textbf{Consistency training weakens right and wrong cues alike, mostly before the answer options, and leaves the cue circuit in use.} (a) Cue following on the same questions, cue before the options. (b) Harmful-cue following removed at each cue position. (c) Share of cue errors corrected by patching, which resets the circuit heads to uncued values. If CT had removed the circuit, the after-CT bar would drop toward the black lines, which show the best random heads (highest over models and methods). All bars and the lines in (b) average the two models. Bars in (a) and the after-CT bar in (c) also average the four methods. Whiskers show ranges, and \llama{} uses one seed in (c).}
\label{fig:hero}
\end{figure*}

Language models often change their answer because of a cue in the prompt. By a \emph{cue} we mean text that points to one answer option without giving valid evidence for it. Examples are the hint ``I believe the answer is B'', an earlier answer in the conversation and few-shot examples labeled with the cued option. Models follow such cues (\emph{cue following}) even when the cued answer is wrong, and they often do not mention the cue when they explain their answer \citep{sharma2024sycophancy,turpin2023language,chua2024bct}.

Consistency training (CT) is a direct remedy. It trains the model to give the same answer with and without the cue. Four CT methods enforce this at different points in the model. Bias-augmented CT (BCT) matches the output tokens \citep{chua2024bct}, activation CT (ACT) matches residual-stream activations \citep{irpan2025consistency} and attention CT (AttCT) matches per-head attention patterns \citep{gautam2026stack}. Rate-matching CT (RMCT) uses reinforcement learning so that the model gives the cued letter about as often with the cue as without it \citep{imran2026ratematching}. Each method comes from its own paper, with its own data, models and tests, and no study has compared all four on equal terms. Two of these papers looked inside the trained models and concluded that BCT learns a different solution from the activation-level methods \citep{irpan2025consistency,gautam2026stack}.

We call the change that CT makes the \emph{fix}. What does the fix actually teach the model? Cues are sometimes right. A good fix should therefore keep the model following a cue that names the correct answer while making it resist one that names a wrong answer. We call such a fix \emph{selective}. A good fix should also hold when the cue moves to a new place in the prompt. The source papers test neither property, because their evaluations use only cues that name a wrong answer and never move the trained cue. We therefore ask three questions. Is the fix selective? Does it hold when the cue moves? Does CT remove the mechanism that carries the cue to the answer, or leave it in use?

We train all four methods with published code, on the same data and on 2 models. We wrote down each prediction before running its experiment \citep{van-miltenburg-etal-2021-preregistering}. The \emph{original} model is the instruction-tuned model before CT. For the third question we use the reader--writer circuit of \citet{gupta2026many}, a few attention heads that carry cues to the answer and already exist after pretraining. Its reader heads attend from the answer position to the cued letter, and its writer heads carry much of the cue's effect on the answer. With it we can test whether the errors left after CT still pass through these heads, and in which layers the fix sits.

\paragraph{Findings.} All four methods make the models follow the trained cue less (\S\ref{sec:efficacy}), and their fixes share three properties (Figure~\ref{fig:hero}). (1) The fix is not selective (\S\ref{sec:content}). A cue that names the correct answer loses about as much influence as one that names a wrong answer, even on the same questions. (2) The fix depends on where the cue sits (\S\ref{sec:position}). It is usually large when the cue comes before the answer options, as in training, and smaller after them. \llama{} RMCT is affected least. (3) The circuit stays in use (\S\ref{sec:mechanism}). Its heads still carry most of the remaining cue-induced errors, and for ACT and AttCT the fix sits mainly in the layers before the circuit. Changing what BCT trains on does not make the fix selective either (\S\ref{sec:variants}). Targets that take the cue's correctness into account mostly make the whole fix weaker. Training with cues at several positions, as in the BCT recipe of \citet{chua2024bct}, changes where the fix works. These results suggest that CT teaches models to give the trained cue less weight without teaching them to judge whether the cue is right.

\paragraph{Implications.} Every CT objective is minimized by ignoring the cue, so we expect correct cues to lose some weight (\S\ref{sec:content}). Evaluations that use only wrong cues at the trained position show neither this loss nor the position effect. We therefore suggest that CT evaluations also test correct cues and new cue positions. We also report two evaluation pitfalls that may affect other work. First, the answer read from the first output token disagreed with the generated answer on up to a fifth of prompts (cf.\ \citealp{wang-etal-2024-answer-c}). Second, merging LoRA adapters into bf16 weights changed the results of \vMergeNFail{} of \vMergeN{} checkpoints by more than a tolerance we set in advance (Appendix~\ref{app:corrections}).

\section{Related work}

\paragraph{Cue following and its mechanism.} Language models repeat back users' stated views \citep{perez2022discovering}, follow stated beliefs and other prompt cues against their own answers \citep{sharma2024sycophancy,wei2023simple}, and often change answers without mentioning the cue \citep{turpin2023language,chua2024bct}. Mechanistic studies locate this behavior in small sets of attention heads \citep{genadi2026sycophancy,sun2026persuaded,pandey2026sharedcircuit}. \citet{gupta2026many} report that the reader--writer circuit is already present after pretraining and that post-training changes when it is used and how strongly it acts. \citet{gupta2026directions} find that alignment tuning largely shapes decodable and steerable directions for such biases. We use the circuit only as a measurement tool, and to our knowledge no study has used it to examine a fix for cue following. The closest results show that alignment training ``masks but does not remove'' a related circuit \citep{pandey2026sharedcircuit} and that DPO bypasses the toxicity mechanism without removing it \citep{lee2024mechanistic}.

\paragraph{Comparisons of CT methods.} \citet{irpan2025consistency} compare training curves on one model and report evidence that BCT and ACT learn ``mechanistically different solutions''. \citet{gautam2026stack} study one checkpoint per method and find that the activation-level methods share one linear pathway while BCT learns a different direction. Other work extends the family with on-policy CT \citep{han2026opct}, CT against political manipulation \citep{phan2026political} and ACT for reasoning models \citep{shah2026act}. \citet{africa2026entrench} compare BCT, ACT and other variants on model organisms that are already sycophantic and find that CT made them more sycophantic, significantly so for ACT. We run all four on the same data and add the self-training control of \citet{chua2024bct} (training without the cue).

\paragraph{Gaps in CT evaluations.} The evaluations of the four methods use only cues that name a wrong answer, and they keep the trained cue where it was in training. Other anti-sycophancy methods have been tested on cues that name the truth. Training a model only to resist persuasion makes it worse at accepting useful persuasion \citep{stengel-eskin-etal-2025-teaching}. A benchmark measures whether deployed assistants accept valid corrections and reject wrong ones \citep{sinha2026sycobench}. Suppressing sycophancy can also make models worse at updating their answer for good reasons when a user pushes back \citep{ma2026suppression,yang2026resist}. Closest to our setting, \citet{tang2026unlearning} train two 32B reasoning models with a target equivalent to BCT's and find that these models agree less with a user who suggests the correct answer. On position, the BCT recipe varies the cue position during training \citep{chua2024bct}. The authors of ACT note that ACT's loss covers only the tokens after the cue, which makes ACT weaker against jailbreaks that add tokens at the end \citep{irpan2025consistency}. A BCT model trained against frustration helps against a persona attack when the facts come before a question but hurts when they come after it \citep[App.~E.6]{gautam2026stack}. Finally, benign fine-tuning can erode safety training \citep{qi2024finetuning}. For consistency-style objectives, we know of only one test of such erosion, for a target that removes deception \citep{tang2026unlearning}. We test correct cues, new cue positions and benign fine-tuning.

\section{Setup}
\label{sec:setup}

\paragraph{Data.} We use \gemmafull{} \citep[\gemma;][]{gemma2_2024} and \llamafull{} \citep[\llama;][]{grattafiori2024llama}. The questions are multiple-choice questions from MMLU, HellaSwag, LogiQA and TruthfulQA \citep{hendrycks2021mmlu,zellers2019hellaswag,liu2020logiqa,lin2022truthfulqa} in the cued form released by \citet{chua2024bct}. We ask for a direct letter answer (Figure~\ref{fig:prompt} in Appendix~\ref{app:instruments} shows an example). Training uses one cue type, a suggested answer (the \emph{trained cue}), on 1,213 questions. ACT's loss covers only the tokens after the cue. We therefore place every training cue before the question, so that this loss covers all of it. The data of \citeauthor{chua2024bct} instead vary the position. As in the BCT recipe, some training cues name the correct answer (\vTrainCorrectCue{} of the 1,213). The \emph{test set} has \vGemNItems{} items, each pairing one question with one cue type, drawn from \vGemNQuestions{} questions that do not appear in training. It contains the trained cue and five held-out cue types. These are a planted earlier answer, a distractor fact, wrong few-shot examples, a distractor argument and spurious few-shot squares. It keeps the cue positions of the source data, which vary (Appendix~\ref{app:details} describes the cues and gives full details).

\paragraph{Methods.} We run all four methods with published code and train LoRA adapters \citep{hu2022lora} on the same prompts (Appendix~\ref{app:details}). BCT fine-tunes the model on cued prompts, with its own uncued answers as targets. It mixes these examples 1:1 with general instruction data. ACT pulls the residual activations of the cued prompt toward those of the original model on the uncued prompt. AttCT matches the per-head attention over the question between the cued and uncued prompts. RMCT rewards sampled answers so that the model gives the cued letter about as often with the cue as without it. Each method keeps its authors' configuration, so the methods also differ in adapter placement and number of updates. We choose learning rates with the F1 score of \citet{irpan2025consistency}, which combines wrong-cue resistance with uncued accuracy and never rewards following correct cues.

\paragraph{Measures.} We measure how often a cue moves the answer, separately for wrong and correct cues. \emph{Harmful-cue following} (flip$\mid$\knew{} in tables) starts from the items that the model answers correctly without a cue. It is the share of them on which a cue that names a wrong option makes the model give that option. \emph{Helpful-cue following} starts from the items that the model answers wrongly without a cue. It is the share of them on which a cue that names the correct option makes the model answer correctly. The \emph{harmful drop} and the \emph{helpful drop} are how much each falls from the original model to the trained model, on the same items. On the test set, we parse greedy generations to a letter and report harmful-cue following for the trained cue and the mean of this rate over the five held-out cue types. For helpful-cue following and for position, we use a \emph{cue battery} of \vGemGenNItems{} test questions with six cue wordings at four positions. These are before the question (\emph{front}, the trained position), just before the answer options (\emph{pre}), just after them (\emph{post}) and after the answer instruction (\emph{end}). Front and pre come before the options, and post and end after them. On the cue battery we read the first answer token, which we check against generation (Appendix~\ref{app:gates}). We count a fix as selective if its harmful drop is at least 0.05 and its helpful drop is at most half of it, the criterion we registered for the BCT variants on \llama. The \emph{relative reduction} at a position is the share of harmful-cue following that a method removes there, $1-{}$method/original.

\paragraph{Circuit instruments.} We use the circuit in three ways (Appendix~\ref{app:instruments} lists its heads). First, we measure how much the reader heads of \gemma{} in layers 13 and 14 attend to the cue. Second, we \emph{patch} the circuit heads, replacing their activations with their values on the uncued prompt. We count how many of the remaining cue-induced errors this corrects and compare with random heads in the same layers. Third, we run a \emph{layer swap}. We copy the attention weights of the trained model into the original model, one block at a time. There are four blocks, two before the circuit, one for the circuit and one after it. We then measure what share of the trained model's fix each mixed model reproduces.

\paragraph{Predictions and statistics.} Before each experiment, we wrote down our predictions and how we would score them. We kept them in a file that we only ever added to, so that these tests do not depend on choices made after seeing the results. Appendix~\ref{app:prereg} reports all of them, including those that failed. Analyses that we did not plan this way are marked exploratory. We first saw on \gemma{} that the fix depends on the cue position, so we treat \llama{} as the fresh test of this pattern. Unless noted, results are means over three seeds, with 95\% bootstrap intervals over test questions (Appendix~\ref{app:details}).

\section{Results}

\subsection{All methods reduce following of the trained cue}
\label{sec:efficacy}

\begin{figure}[t]
\centering
\includegraphics[width=\columnwidth]{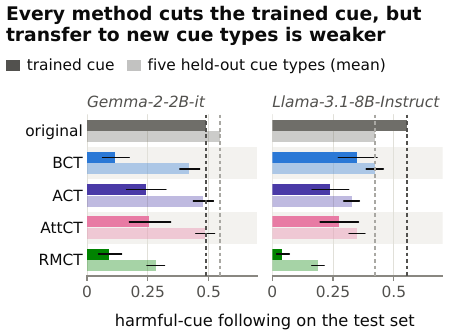}
\caption{\textbf{Every method cuts the trained cue, but transfer to new cue types is weaker.} Whiskers show 95\% intervals and dotted lines the original model (Table~\ref{tab:main}).}
\label{fig:efficacy}
\end{figure}

We first check that each method does its job, which is to make the model resist a cue that names a wrong answer. Figure~\ref{fig:efficacy} shows harmful-cue following on the test set, for the trained cue and for five cue types not used in training. Every method reduces following of the trained cue on both models. On \gemma, BCT and RMCT remove three quarters or more of it, and ACT and AttCT remove about half. Our prediction that every method would at least halve it on \gemma{} \vd{\vVerdictPOne} narrowly. AttCT falls just short of half and ACT just passes it, and for both the threshold lies inside the error bars. On \llama, RMCT removes nearly all of it and BCT removes the least, although the interval for the gap between BCT and AttCT includes zero.

Transfer to the five held-out cue types is weaker. RMCT transfers best on both models. We had predicted that BCT would transfer best on \gemma, but it does not, and on \llama{} it barely changes following of held-out cues. On \gemma, much of RMCT's advantage comes from a preference for one answer letter. It answers D on \vGemRmctModalShare{} of uncued questions, and a model that leans toward one letter can follow cues that name other letters less often. As a control, we shift the original model's answer-letter frequencies to roughly match those of RMCT. Against this shifted model, RMCT's drop on held-out cues is close to that of BCT (exploratory, Table~\ref{tab:letterprior}). RMCT keeps a clear lead on \llama{} under this control. Training costs little uncued accuracy on the test set (at most 0.05 for every method), but \gemma{} BCT loses accuracy on the AttCT authors' MMLU evaluation (seed 42, Appendix~\ref{app:authors}).

\subsection{The fix is not selective}
\label{sec:content}

\begin{figure}[t]
\centering
\includegraphics[width=\columnwidth]{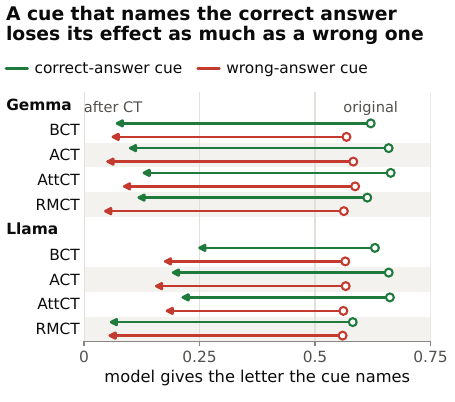}
\caption{\textbf{A cue that names the correct answer loses its effect as much as a wrong one.} Arrows run from the original model (circle) to the model after CT, on questions that both models answer wrongly without a cue (cue before the options). A selective fix would keep the green arrows short.}
\label{fig:sameq}
\end{figure}

Does a model that resists wrong cues still follow correct ones? We test this in two ways.

First, we hold the question fixed, with the cue before the options (exploratory). On questions that the model answers wrongly without a cue, we give either a cue that names the correct answer or a cue that names another wrong option. Figure~\ref{fig:sameq} shows how often the model gives the cued letter before and after training. Before CT, the model follows each cue on more than half of these questions. After CT, it follows both far less often. The green and red arrows are about equally long, so the two cues lose about the same influence. If anything, the correct cue loses more in absolute terms. Across the four methods, the correct cue's drop minus the wrong cue's drop ranges from \vSameDiffAllMin{} to \vSameDiffAllMax{} (Table~\ref{tab:sameitem}).

Second, we compare the helpful and harmful drops pooled over positions (exploratory, Table~\ref{tab:content}). For a selective fix, the helpful drop would be at most half of the harmful drop. Instead, the helpful drop is \vGemRmctRatio--\vGemAttctRatio{} times the harmful drop on \gemma{} and \vLlaRmctRatio--\vLlaAttctRatio{} times on \llama. (\llama{} BCT has a ratio of \vLlaBctRatio{} because it barely lowers harmful-cue following.) Other scales agree. Before the options, where the fix is strong, the ratio is at least \vLlaBeforeLogMin{} whether we measure absolute drops, proportional drops or log-odds shifts (Tables~\ref{tab:scale} and~\ref{tab:contentbypos}). Two cases differ. After the options, where the fix is weak, the helpful drop is mostly the smaller one in absolute and proportional terms. In the same-question test, \gemma{} RMCT keeps little cue following, and on the log-odds scale this remainder favors correct cues more than before (Appendix~\ref{app:selectivity}). Our prediction that every method lowers helpful-cue following in every seed holds on both models.

Why would CT suppress correct cues? Each CT objective is minimized by ignoring the cue, and the training cue names a wrong option in 80\% of the training questions. Even when the cue names the correct answer, the BCT target (the model's own uncued answer) often disagrees with the cue (on \vGemBctContra{} of the \vTrainCorrectCue{} such questions for \gemma{} and \vLlaBctContra{} for \llama). What is new is the size of this cost, and that all four methods share it.

\subsection{The fix depends on the cue position}
\label{sec:position}

\begin{figure}[t]
\centering
\includegraphics[width=\columnwidth]{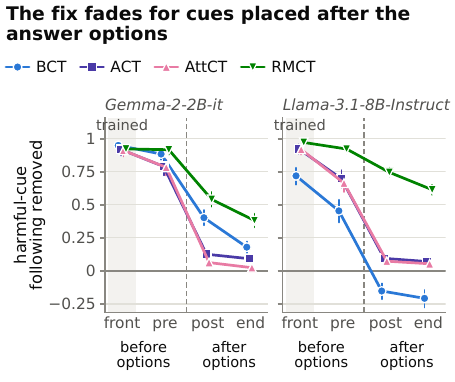}
\caption{\textbf{The fix fades for cues placed after the answer options.} Whiskers show 95\% intervals. Training puts every cue at the front (shaded). On \gemma, the values of RMCT after the options come mostly from its preference for answering \vGemRmctModalLetter{} (\S\ref{sec:position}).}
\label{fig:position}
\end{figure}

Training places every cue before the question. Does the fix still hold when the cue appears elsewhere? We place the same cues at four positions and measure the share of harmful-cue following that each method removes (Figure~\ref{fig:position}, Table~\ref{tab:content}). The fix works well before the answer options and fades after them. Before the options, the reductions are large for every method, except for \llama{} BCT at the pre position. At the end of the prompt, ACT and AttCT remove less than a tenth of harmful-cue following on both models, and \llama{} BCT even makes it worse than in the original model.

\llama{} RMCT is the exception, because its fix depends much less on the position. Its reduction at the front exceeds its reduction just after the options by only \vLlaRmctFrontMinusPost{} (95\% \vLlaRmctFrontMinusPostCI), so our prediction that the reduction of every method falls by at least 0.30 there \vd{\vVerdictLThree}. \gemma{} RMCT also looks less position-dependent, but this rests mostly on its preference for \vGemRmctModalLetter{} (\S\ref{sec:efficacy}). A model that answers \vGemRmctModalLetter{} whatever the cue says looks as if it resists every cue that names another letter. We therefore add a stricter count. Every wrong answer counts as following the cue, and we leave out the questions whose correct answer is the letter that the checkpoint most often gives without a cue. By this count, \gemma{} RMCT removes only \vGemRmctNCxPost{} just after the options and \vGemRmctNCxEnd{} at the end, no more than BCT (\vGemBctNCxPost{} and \vGemBctNCxEnd, exploratory, Appendix~\ref{app:position}).

Are cues after the options simply stronger to begin with? To check, we compare only combinations of wording and position that the original model follows at similar rates before and after the options (exploratory, Table~\ref{tab:poscheck}). ACT on \llama{} still removes \vLlaActMatchBefore{} of harmful-cue following before the options and only \vLlaActMatchAfter{} after them. Before the options, following falls about as much for novel wordings as for wordings from the training family. So position, more than strength or wording, seems to decide how well the fix works.

The test set shows the same pattern. Two thirds of its trained-cue items have the cue at a position other than the front (exploratory, Table~\ref{tab:layout}). With the cue first, every method removes \vLayFrontMin--\vLayFrontMax{} of harmful-cue following, while after the options ACT and AttCT remove only \vGemAttctLayPost--\vLlaActLayPost. For BCT and RMCT the results vary more. Both the lack of selectivity and the dependence on position also hold at the neighboring learning rates. For \llama{} ACT and AttCT, they also hold on the AttCT authors' own data (exploratory, seed 42, Table~\ref{tab:lrsweep}).

\subsection{The circuit stays in use}
\label{sec:mechanism}

\begin{figure*}[t]
\centering
\includegraphics[width=\textwidth]{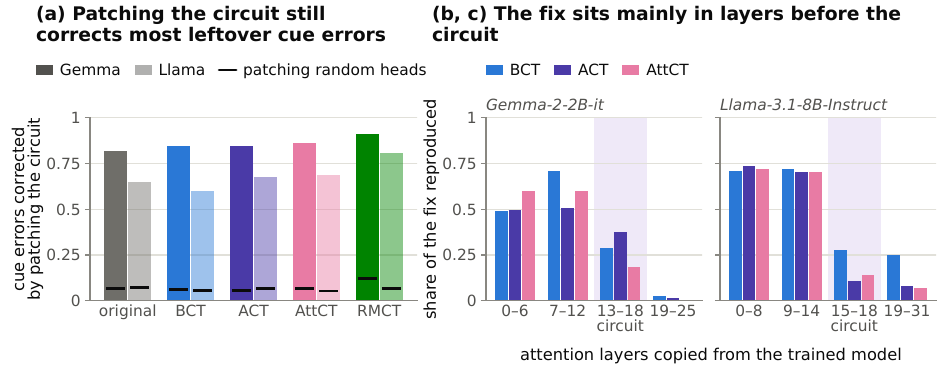}
\caption{\textbf{The circuit still carries the leftover errors, and the fix sits mainly before it.} (a) Share of the remaining cue-induced errors that patching the circuit heads corrects. (b, c) We copy one block of attention layers from the trained model into the original model and show the share of the method's fixes that this mixed model reproduces (our planned readout for \llama). The swap cannot test RMCT. \llama{} uses one seed.}
\label{fig:circuit}
\end{figure*}

Does CT remove the mechanism that carries the cue, or only weaken its effect? If CT had removed the circuit, the errors that remain after training would no longer pass through it. Figure~\ref{fig:circuit}a shows that they still do. For every method, patching the circuit heads corrects most of the remaining cue-induced errors (\vLlaBctPatch--\vGemRmctPatch), much as in the original model. Random heads in the same layers correct at most \vGemRmctPatchRandom{} (Table~\ref{tab:patch}). This was our prediction for \llama, and it \vd{\vVerdictLFour}. Before and after CT, most of these errors are on held-out cues.

If the circuit stays in use, where does the fix sit? The layer swap answers this (Figure~\ref{fig:circuit}b, c, Table~\ref{tab:route}). For ACT and AttCT, the fix sits mainly in the layers before the circuit. Copying either block before the circuit reproduces \vRouteBelowMin--\vRouteBelowMax{} of the trained model's fixes on both models. The layers after the circuit reproduce at most \vRouteAboveMax. \gemma{} BCT behaves the same way. For \llama{} BCT, however, the layers after the circuit reproduce a quarter of its fixes, so our prediction of under 15\% fails, although 15\% lies inside the interval. The swap copies only attention weights, and RMCT also trains MLP and output weights, so this test cannot locate its fix (Appendix~\ref{app:mechanism}).

The attention of the \gemma{} reader heads fits this picture for ACT and AttCT (Table~\ref{tab:reading}). On the items these methods fix, the reader heads attend about a fifth less to the trained cue. On the items where the cue still works, attention falls by \vGemActReadStill{} or less. Across all labeled trained-cue items (Appendix~\ref{app:instruments}), it falls by only 11--14\% for ACT and AttCT, less than the 30\% we predicted. So on the trained cue, ACT and AttCT may fix an item partly by reducing what reaches the reader heads, though this evidence is only correlational.

\subsection{Changing the training data}
\label{sec:variants}

\begin{figure}[t]
\centering
\includegraphics[width=\columnwidth]{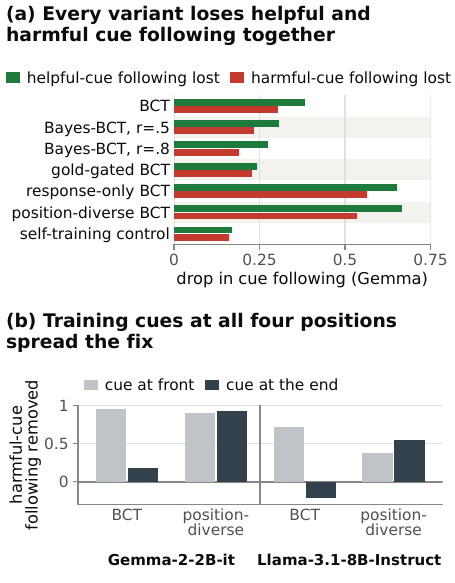}
\caption{\textbf{No change to the training data of BCT makes the fix selective.} Cue battery, seed means. (a) shows each BCT variant and the self-training control on \gemma, and (b) compares BCT with position-diverse BCT, which trains with the cue at all four positions.}
\label{fig:variants}
\end{figure}

If CT is not selective because its targets ignore whether the cue is right, then targets that take the cue's correctness into account should help. We test two such targets for BCT, and two changes to what BCT trains on (Figure~\ref{fig:variants}, Table~\ref{tab:variants}).

\emph{Bayes-BCT} samples each target from the model's uncued answer distribution after shifting it toward the cued letter, as if the cue were right with probability $r$. \emph{Gold-gated} BCT uses the correct label. It keeps a target that is right, corrects a wrong target when the cue is right, and drops the rest. Neither becomes selective. Figure~\ref{fig:variants}a shows that both lose less helpful-cue following than BCT, but they also remove less harmful-cue following, so the two drops shrink together. With $r{=}0.8$ on \gemma, the ratio of Bayes-BCT is no better than that of BCT (\vGemBayesEightCAI{} against \vGemBctCAI), and gold-gated BCT stays well above one half (\vGemGoldCAI). We had predicted that on \gemma{} each would keep at least half of the harmful drop of BCT with at most half of its helpful drop. This prediction fails for both Bayes-BCT ($r{=}0.8$) and gold-gated BCT. On \llama, neither variant is selective, mostly because, pooled over positions, neither clearly lowers harmful-cue following.

The other two changes alter what BCT trains on. They help with position but leave the fix unselective. \emph{Position-diverse} BCT places each training cue at one of the four positions, as the BCT recipe of \citet{chua2024bct} does. Its fix then also works at the end position on both models (Figure~\ref{fig:variants}b), so both of our predictions for it hold. On \llama, however, it halves the fix when the cue comes first, and its helpful drop is the largest of the variants. \emph{Response-only BCT} \citep{irpan2025consistency} masks the prompt from the loss, drops the instruction data, and its selected learning rate is ten times larger. It is the strongest \gemma{} method on held-out cues, and it depends much less on position even though every training cue comes first. It is still just as unselective (ratio \vGemBctroCAI), and its MMLU accuracy falls slightly below the 0.55 we predicted (Appendix~\ref{app:authors}).

The \emph{self-training control} of \citet{chua2024bct} measures how much of the effect of BCT comes from fine-tuning on the model's own answers alone. It uses the same targets and instruction data without the cue. It increases cue following on \llama. On \gemma, it achieves two thirds of the reduction of BCT for the trained cue on the test set. On the cue battery it reaches about half of the harmful drop of BCT and just under half of its helpful drop (Table~\ref{tab:variants}). So on \gemma, fine-tuning on its own answers explains much of what BCT does.

\subsection{Relapse after benign fine-tuning}
\label{sec:relapse}

\begin{figure}[t]
\centering
\includegraphics[width=\columnwidth]{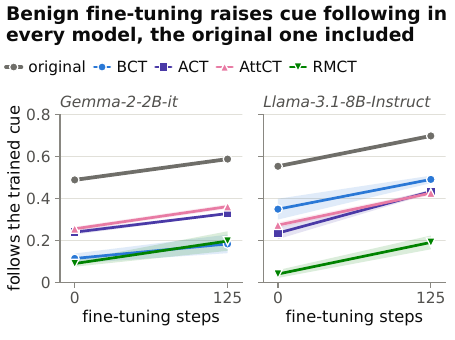}
\caption{\textbf{Benign fine-tuning raises cue following in every model, the original one included.} Harmful-cue following on the trained cue, with bands for the SD over seeds (the original model is one run).}
\label{fig:relapse}
\end{figure}

Finally, does the fix survive later benign fine-tuning, which can erode safety training \citep{qi2024finetuning}? We fine-tune each trained model for 125 steps on benign instruction data (Figure~\ref{fig:relapse}, Table~\ref{tab:relapse}). Cue following rises for every method, but it also rises in the original model, so part of the apparent relapse seems to be a general effect of the fine-tuning. We planned to measure relapse as the share of each method's improvement that is lost. By this measure, the \gemma{} methods lose \vGemBctRelFracU--\vGemAttctRelFracU{} of it. Our prediction that every method loses at least a quarter \vd{\vVerdictROne} because of BCT (\vGemBctRelFracU, 95\% interval over seeds \vGemBctRelFracSeedCI). After we subtract the original model's own increase, every \gemma{} method is within seed noise of zero. On \llama, ACT has the largest such value (\vLlaActRelAdjU). It is positive in every seed, but its seed interval includes zero. On the log-odds scale, however, RMCT rises more than ACT (Appendix~\ref{app:relapse}). On held-out cues, cue following rises more for some methods than for the original model (Table~\ref{tab:relapse}, without the few-shot correction of Appendix~\ref{app:fewshot}).

\section{Discussion}

\paragraph{Predictions.} For our three main findings, the predictions that every method lowers helpful-cue following and that the circuit still carries the remaining errors both hold. Two others fail because of one method each. The fix of \llama{} RMCT fades after the options by less than we predicted. For \llama{} BCT, the layers after the circuit reproduce more of the fix than the 15\% ceiling we predicted. The clearest failure concerns the \gemma{} reader heads. Their attention falls by much less than we predicted (\S\ref{sec:mechanism}), and for BCT on some held-out cues it changes more than we expected. In total, 14 of the 27 scored outcomes hold (Appendix~\ref{app:prereg} lists all of them). Several pass or fail within the error bars, and one passes narrowly, since BCT lowers following of held-out cues on \llama{} by only 0.0005.

\paragraph{One pattern across four methods.} Although the four methods act on different parts of the model, their fixes look alike. The account most consistent with our results is that the reader--writer circuit already exists after pretraining \citep{gupta2026many} and that all four methods turn down how strongly it moves the answer. ACT and AttCT seem to do so partly by changing what reaches the reader heads. \citet{gupta2026grounding} report a similar pattern for context grounding, where gains from SFT and DPO largely reuse machinery already in the starting model. The earlier evidence that BCT learns a different solution comes from training curves and linear directions. These do not measure this circuit, so the two findings need not conflict. The methods differ in how strong the fix is, how far it transfers and, for \llama{} RMCT, how much it depends on position, but all treat right and wrong cues alike.

\paragraph{What CT evaluations should measure.} Our rule for selecting learning rates, like the source evaluations, never rewards following a correct cue. We suggest that CT evaluations also test correct cues and cues at new positions, on the same questions as the wrong cues. They should also state whether ``held-out'' means new questions or new cue types (Appendix~\ref{app:authors}). Objectives that do not reward ignoring a correct cue may keep correct cues effective. \citet{wang2026eit} train with reinforcement learning on cues that name the correct answer half the time, and reward correct answers under both kinds of cue. Their models keep performance when the cue is right. Preference optimization can teach a model when to accept persuasion \citep{stengel-eskin-etal-2025-teaching}, and in one benchmark, supervised training on the same targets does not \citep{yang2026switch}. Of the four methods, RMCT transfers furthest, partly through its letter preference on \gemma. It also needs 256 samples per question. No method gives a model that weighs cues by their merit.

\section{Conclusion}

Trained on the same data, four consistency-training methods behave alike. Each makes the cue move the answer less whether the cue is right or wrong, and does so mostly when the cue comes before the answer options, as in training. None teaches a model when to trust a cue or removes the circuit that carries it. Supplementary code is available at: \url{https://github.com/prakharg55/bias_ct_eval}

\section*{Limitations}

We note a few limitations.
\begin{list}{(\roman{limctr})}{\usecounter{limctr}\setlength{\leftmargin}{1.7em}\setlength{\labelwidth}{1.5em}\setlength{\labelsep}{0.3em}\setlength{\itemsep}{1pt}\setlength{\topsep}{2pt}\setlength{\parsep}{0pt}}
\item We study one setting on two models (2B and 8B) trained with LoRA. The models give direct letter answers to multiple-choice questions, and training uses one cue placed first (except in position-diverse BCT). BCT and RMCT were designed for chain-of-thought answers, and our RMCT runs use the authors' algorithm but not their training setup (Appendix~\ref{app:authors}).
\item Each method keeps its authors' configuration, so the methods also differ in adapters, batch size and number of updates. Our runs also depart from the source recipes in ways listed in Appendix~\ref{app:details}. Rankings between methods may reflect these choices.
\item Learning rates are selected on one seed, and the \llama{} patching and layer-swap results also use one seed. The self-training control and the BCT variants other than response-only BCT reuse the learning rate of BCT. Several predictions pass or fail within the error bars, some analyses are exploratory and post hoc, and the attention comparison is correlational.
\item The benign fine-tuning data share a source file with the instruction data of BCT, and relapse beyond the original model's own increase has wide intervals.
\end{list}

\section*{Ethical considerations}

We use only public models, datasets and code, and we involve no human subjects. Fixes that make a model ignore wrong user cues can also make it ignore correct ones, so we suggest testing for this before deployment.

\paragraph{Use of AI assistants.} The authors used generative AI tools for experiments, verification of results, and language editing. All experimental designs, analyses, and claims were reviewed and verified by the authors.

\section*{Acknowledgments}
This material is based in part upon work supported
by the German Federal Ministry of Education and Research (BMBF): Tübingen AI Center, FKZ: 01IS18039B; by the Machine Learning Cluster of Excellence, EXC number 2064/1 – Project number 390727645; % if Zhijing has the MPI or Tuebingen affiliation
% by NSERC Discovery Grant RGPIN-2025-06491; % for causality research
% by the University of Toronto’s Acceleration Consortium, which receives funding from the Canada First Research Excellence Fund (CFREF); % causality; also AI for science work
% by the Frontier Model Forum and AI Safety Fund; % for multi-agent and sanctioning research
by Coefficient Giving; % if the work is general AI safety-related
by Schmidt Sciences; % if the worker is on the topic of Mech Interp or test data contamination
% by the Canadian AI Safety Institute Research Program at CIFAR through a Catalyst Award; % multicultural awareness for LLMs
% by the Survival and Flourishing Fund; % democraccy work
by the Cooperative AI Foundation; by the Canadian AI Safety Institute Research Program at CIFAR. % for multi agent research
Resources used in preparing this research project were provided, in part, by the Digital Research Alliance of Canada; the Province of Ontario; the Swiss AI Compute grants; the Government of Canada through CIFAR; and companies sponsoring the Vector Institute.
We thank Jiarui Liu, Andrew Kim, and the other key developers of PaperMentor \citep{liu2026papermentorhumancenteredmultiagentwriting}, which provides mentorship and valuable suggestions in polishing this paper after the first draft by the authors. We further acknowledge the crucial work of AdminBot for administrative support that streamlined our research workflow.

\bibliography{references}

\appendix
\raggedbottom
% let up to four full-width tables share the top of an appendix page
\setcounter{dbltopnumber}{4}\renewcommand\dbltopfraction{0.9}\renewcommand\dblfloatpagefraction{0.8}   % appendix: no stretched paragraph gaps above floats

\section{Pre-registration}
\label{app:prereg}
We appended every prediction to an append-only file before running the experiments it concerns. The supplementary material contains a copy in which only the names of the compute cluster and of a queue are redacted. The timestamps are those written in the file, and we cannot certify them externally. We wrote each block of predictions after the results of earlier blocks were known. P1--P5 followed one-epoch pilot runs at the reference learning rates on seed 42. The 27 Sep block followed the \gemma{} results of P1--P5, including the position pattern. The 30 Sep block followed the results of the 27 Sep block, and S1--S2 followed the merged results of all other predictions. The earlier sections of the file record a retired design (our own re-implementations of the objectives), whose predictions we do not score. Analyses without a registered prediction are exploratory. They include the selectivity ratios of the four methods and their scales, the same-question test, the position and strength splits, the letter-prior control, the reading split by fixed items, the \gemma{} patching and layer swap, the head subsets, and the drift-adjusted and log-odds relapse.

\subsection{Registered predictions}
\begin{table*}[t]\centering\scriptsize\setlength{\tabcolsep}{3pt}
\begin{tabular}{l l p{11.5cm} l}
\toprule
code & registered (EDT) & prediction, as written & outcome \\
\midrule
P1 & 26 Sep, 23:47 & efficacy: every arm cuts trained-cue (suggested\_answer test) flip$\mid$knew by $\geq$ 50\% vs untrained. & \vd{\vVerdictPOne} \\
P2 & 26 Sep, 23:47 & transfer: BCT has the lowest mean held-out caving (5 held-out cues) of the four arms. & \vd{\vVerdictPTwo} \\
P3 & 26 Sep, 23:47 & reading vs following (wave-3 W2 pattern): early-reader cue-span mass on the trained cue falls by $\geq$ 30\% for ACT and AttCT in all seeds, while on every held-out cue each arm's relative change stays within $\pm$25\% (seed mean). & \vd{\vVerdictPThree} \\
P4 & 26 Sep, 23:47 & where edits land: per-head OV depth centroid (log 82.7 definition) is lower for AttCT and ACT than for RMCT in all seeds. & \vd{\vVerdictPFour} \\
P5 & 26 Sep, 23:47 & content-blindness: every arm lowers paired helpful-cue uptake (gen3, W4c paired definition) in all seeds. & \vd{\vVerdictPFive} \\
\midrule
F1 & 27 Sep, 14:21 & Bayes-BCT(0.8): helpful-uptake drop $\leq 0.5\times$ BCT's AND harmful-caving drop $\geq 0.5\times$ BCT's. & \vd{\vVerdictFOne} \\
F2 & 27 Sep, 14:21 & Gold-gated BCT: the same two conditions. & \vd{\vVerdictFTwo} \\
F3 & 27 Sep, 14:21 & Position-diverse BCT: relative reduction of harmful caving at the END position $\geq 2\times$ BCT's (BCT: 18\%). & \vd{\vVerdictFThree} \\
R1 & 27 Sep, 14:21 & every CT method's trained-cue relapse fraction at step 125 is $\geq$ 0.25 (seed mean). & \vd{\vVerdictROne} \\
L1 & 27 Sep, 14:21 & every arm lowers trained-cue flip$\mid$knew and held-out caving vs untrained (seed means). & \vd{\vVerdictLOne} \\
L2 & 27 Sep, 14:21 & content-blind: every arm's paired helpful-uptake drop $>$ 0 in every seed. & \vd{\vVerdictLTwo} \\
L3 & 27 Sep, 14:21 & position-bound: for every arm, the relative reduction of harmful caving at post AND at end is $\geq$ 30 points below that at front (seed means). & \vd{\vVerdictLThree} \\
L4 & 27 Sep, 14:21 & circuit intact: for every arm (seed 42), circuit patch recovery $\geq 0.8\times$ the untrained model's and $\geq 3\times$ the best random set's. & \vd{\vVerdictLFour} \\
L5 & 27 Sep, 14:21 & RMCT (anchor 0) is degenerate (Amendment 2) at its published LR 2.86e-4. & \vd{\vVerdictLFive} \\
\midrule
LF1 & 30 Sep, 08:50 & Bayes-BCT(0.8) is NOT content-aware, and its paired harmful drop is $\leq$ Llama BCT's (.038) (its targets follow wrong hints on 25\% of known items, so it trains cue-following). & \vd{\vVerdictLFOne} \\
LF2 & 30 Sep, 08:50 & Gold-gated BCT is NOT content-aware (as on gemma, where helpful and harmful drops were equal). & \vd{\vVerdictLFTwo} \\
LF3 & 30 Sep, 08:50 & Position-diverse BCT: relative reduction of harmful caving at the END position $\geq$ .50 (Llama BCT: $-$.22) AND helpful uptake at the END position falls by $\geq$ .30 absolute vs untrained (.858) -- the muting generalises, content-blind. & \vd{\vVerdictLFThree} \\
LR1 & 30 Sep, 08:50 & ACT and AttCT: trained-cue relapse fraction at step 125 $\geq$ .25 (seed means). & \vd{\vVerdictLROne} \\
LR2 & 30 Sep, 08:50 & BCT's trained-cue relapse fraction is below ACT's and below AttCT's. & \vd{\vVerdictLRTwo} \\
B1 & 30 Sep, 08:50 & content-blind: paired helpful-uptake drop $>$ 0 in every seed. & \vd{\vVerdictBOne} \\
B2 & 30 Sep, 08:50 & position-bound: relative reduction of harmful caving at front minus that at end $\geq$ 30 points (seed mean). & \vd{\vVerdictBTwo} \\
B3 & 30 Sep, 08:50 & less accuracy cost than BCT: MMLU clean accuracy on the AttCT evaluator $\geq$ .55 (BCT .512, untrained .596), seed 42. & \vd{\vVerdictBThree} \\
D1 & 30 Sep, 08:50 & for each of BCT, ACT, AttCT the single-scope weight splice with the largest forward mediated fraction is EARLY or CARRIER (below the circuit), and LATE mediates $<$ .15. & \vd{\vVerdictDOne} \\
\midrule
S1 & 1 Oct, 19:11 & the self-training control's cut in trained-cue flip$\mid$knew vs untrained is less than half of BCT's cut, on each model. & \vd{\vVerdictSOneGem}\,/\,\vd{\vVerdictSOneLla} \\
S2 & 1 Oct, 19:11 & its paired helpful-uptake drop (gen battery, W4c paired definition) is less than half of BCT's, on each model. & \vd{\vVerdictSTwoGem}\,/\,\vd{\vVerdictSTwoLla} \\
\bottomrule
\end{tabular}
\caption{Every scored prediction as registered (2026). P1--P5, F1--F3 and R1 concern \gemma, and B1--B3 concern response-only BCT on \gemma. L1--L5, LF1--LF3, LR1--LR2 and D1 concern \llama. S1--S2 outcomes are given as \gemma{}\,/\,\llama. The 30 Sep 12:07 erratum withdrew the composition numbers of the 30 Sep block, including the 25\% in LF1, and the \llama{} Bayes-BCT data for LF1 were rebuilt on the corrected readout. The P1--P5 section header in the file reads 27 Sep, and an erratum appended one minute later corrects it to 26 Sep. Outcomes are on the corrected, unmerged results.}
\label{tab:prereg}
\end{table*}
Table~\ref{tab:prereg} quotes each scored prediction as written, with its registration time and outcome. We typeset inequalities and multiplication and keep internal names such as ``gen3'' and ``W4c'', which the file defines. ``Log 82.7'' points to the project log, whose depth definition is the one in Appendix~\ref{app:instruments}. The reference values that some predictions quote (for example ``BCT: 18\%'') are those known when they were written, on the readouts then in use.

\appflush
\subsection{Amendments}
\begin{table*}[t]\centering\scriptsize\setlength{\tabcolsep}{3pt}
\begin{tabular}{l p{12.7cm}}
\toprule
appended (EDT) & amendment, erratum or note \\
\midrule
27 Sep, 00:27 & LR rule, Amendment 1 (the header says 00:45, corrected by an erratum). The grid-end test uses the best F1, extensions are sequential and at most two, and an untied published LR yields to the best-F1 LR. RMCT execution v2 uses shared-prefix sampling and a one-epoch run in resumable segments (we verified that the LoRA weights and the Adam step are identical across boundaries). \\
27 Sep, 02:07 & Exploratory RMCT+anchor variant declared (anchor weight 0.5, the trainer's default), after the pre-registered RMCT runs had collapsed to a constant answer at every LR evaluated so far. It is not part of P1--P5. \\
27 Sep, 02:31 & LR rule, Amendment 2. A checkpoint that gives one letter on at least 95\% of validation prompts is degenerate and excluded. If every grid point is degenerate, the grid extends downward. \\
30 Sep, 09:01 & Erratum to D1. RMCT cannot be tested by the layer swap (its adapter trains MLP and unembedding weights), so it is dropped from that analysis. \\
30 Sep, 11:04 & \llama{} relapse runs use micro-batches of 2 because of memory. The loss and gradient are unchanged up to bf16 numerics. \\
30 Sep, 12:07--13:27 & \llama{} readout correction, in four notes. The first-token readout must include the ``(L'' tokens, with a readout gate of at least .97 agreement with greedy generation (v6, 12:07, with gate coverage and handling at 12:24). Because v6 fails for \llama{} BCT, the readout becomes the maximum over letter forms (v7), and \llama{} LR selection, test rates and relapse move to generated answers (12:53). Exact bf16 ties are broken as greedy decoding breaks them (v7.1, 13:27). \\
30 Sep, 20:33--1 Oct, 05:13 & Readout gates, in six notes. A cue-battery gate is declared (20:33). It fails for \gemma{} BCT, so every \gemma{} cue-battery checkpoint is re-scored on v7.1 (23:14). The test-set gate is extended to all \gemma{} checkpoints (31 at the time), and failing checkpoints use generated answers, and v7.1 for their forward-pass instruments including patch recovery (01:45, 01:52). At the first failure (BCT seed 44), its letter-prior control and patch recovery are recomputed against an original-model reference on the same readout (04:13). A regression exception records that one item differs for one checkpoint because of an exact bf16 tie (05:13). \\
1 Oct, 06:18 and 07:56 & \gemma{} relapse re-scored on generated answers, because the first-token readout fails its gate after benign fine-tuning. The exploratory RMCT-anchor relapse is re-run on the same basis. \\
1 Oct, 11:19 & Pre-write-up audit. Three checks are declared with their rules, namely adapter-merge fidelity, few-shot cues scored at the test slot, and \gemma{} LR selection on generated answers. \\
1 Oct, 19:11 & Self-training control declared, with S1--S2. \\
1 Oct, 19:43 and 20:11 & Merge-fidelity outcome. All seven checkpoints checked by then fail the declared rule (48 of 51 in the end, Appendix~\ref{app:merge}). Every adapter result, including the layer swap, is re-run unmerged. The RMCT check in the paper's regime stays on the authors' merged pipeline. \\
1 Oct, 21:26 & Outcome of the unmerged LR re-selection, recorded before any re-selected test result was read. The selections for \gemma{} ACT and AttCT change, and every other selection stands. \\
9 Oct, 00:12 and 00:28 & Exploratory analyses declared. The cue battery at the neighboring learning rates of every sweep, and on the \llama{} checkpoints trained with the authors' recipe (Table~\ref{tab:lrsweep}). \\
\bottomrule
\end{tabular}
\caption{Amendments, errata and correction notes after the first scored predictions, in the order appended (2026). Notes on one topic appended within hours of each other share a row. The full texts are in the supplementary pre-registration file.}
\label{tab:amend}
\end{table*}
Table~\ref{tab:amend} lists every later amendment, erratum and correction note in order. Most notes state what had or had not been run when they were written. We appended rule changes before computing in full the outputs they apply to, with exceptions that the notes record. Amendments 1 and 2 were written after the base-grid validation results they apply to. The \llama{} readout notes (12:07, 12:53) followed its seed-42 generated test rates, and the 12:53 note also followed the v6 validation results of the \llama{} sweep. The 23:14 note followed a subsample re-score that had shown changes of at most .02. The 05:13 note accepts an observed discrepancy. P1 and P2 also fail on the superseded summed readout (05:13 note), and the unmerged re-run replaces the other values in the record that use that readout. Appendices~\ref{app:selection} and~\ref{app:merge} give the selection and merge reports that the notes promise. Several smaller side reports that the notes promise are not included, for example the \gemma{} test rates and cue-battery values on the superseded summed readout and the first-position relapse fractions. Re-scoring the cue battery changed no verdict.

\appflush
\subsection{Verdicts with intervals}
\begin{table*}[t]\centering\scriptsize\setlength{\tabcolsep}{3pt}
\tabVerdicts
\caption{Pre-registered verdicts that turn on a single estimate, with the deciding quantity, threshold, value and 95\% interval. The intervals come from a bootstrap over questions for test-set, cue-battery and layer-swap rates, a binomial bootstrap for relapse fractions (Table~\ref{tab:relapse} gives the seed $t$ interval) and a seed-level $t$ interval for LR2 and S1--S2. A verdict is unresolved when the interval contains the threshold.}
\label{tab:verdicts}
\end{table*}
Table~\ref{tab:verdicts} gives the deciding value and interval for the verdicts that turn on a single estimate, and the other verdicts have no computed interval. Several verdicts are decided inside their intervals. These are P1 (ACT, AttCT), L1 (BCT, held-out), R1, LR2, B3, D1 (BCT, ACT), and S1 and S2 on \gemma{} (B3 in Table~\ref{tab:authors}). LF3 holds narrowly on both of its conditions. L1 (BCT, held-out) failed under the superseded readout.

\appflush
\section{Experimental details}
\label{app:details}
\subsection{Data}
The shared training set is rows 100--500 of the suggested-answer cue in the test dumps that \citet{chua2024bct} release for MMLU, HellaSwag, LogiQA and TruthfulQA. The dumps are formatted for chain-of-thought answers and put the cue before the question, before the options, after the options or after the answer instruction. Our loader replaces the reasoning instruction with a direct-answer one, which also removes the cues placed after that instruction, so we drop the pairs whose cued and uncued prompts become identical (24\% of rows). This leaves 1,213 questions. We move every training cue before the question (the cue, a newline and the uncued prompt), where the loss of ACT over the longest shared token suffix covers the whole question. The cues use 51 templates of the suggested-answer family. In 6 rows an extraction error cuts the first letter of the cue and adds a line with a stray ``A''. \vTrainCorrectCue{} of the cues name the correct answer, and we keep them, as in the BCT and AttCT recipes.

The BCT targets are each model's own temperature-1 answers to the uncued prompts, produced by the generator of the AttCT authors (\gemma{} 45\% correct, \llama{} 53\%), with the \gemma{} targets re-rendered in its chat template. As in \citet{chua2024bct}, we mix them 1:1 with rows of the instruction file of the AttCT repository, whose responses come from other models (gold-gated BCT keeps all 1,213 instruction rows).

The validation set is rows 500--600 of the trained cue (LogiQA 500--549). It has 265 items, 11 of which share questions with the test set (distractor argument, LogiQA). The test set is rows 1200--1300 of six cue types, or the last 100 rows where a dump has fewer than 1,250 usable rows (LogiQA and TruthfulQA, whose dumps are shorter, and the distractor argument, whose dumps lack many cued prompts). Besides the trained cue, these are a planted earlier answer (an earlier assistant turn that gives the cued letter), a distractor fact about the cued option, wrong few-shot examples (the test question appears as an example with the cued label), a distractor argument, and spurious few-shot squares (a square marks the labeled answer in each example and the cued option in the test question). It shares no questions with training and has \vGemNItems{} items (\vLayNTotal{} for the trained cue) over \vGemNQuestions{} questions, with the cues at the original positions of the dumps. Our planted-answer (``post hoc'') cue differs from that of \citeauthor{chua2024bct}, because the loader drops their sentence ``Don't anchor on your previous response''. Our cue is therefore a planted earlier commitment.

\subsection{Training}
BCT, ACT and AttCT run through the AttCT repository, unmodified at commit b3c1896, with copies of its configurations. For ACT and AttCT, a dataloader shim feeds the shared prompts. RMCT runs the authors' trainer verbatim at commit 4392d15 of their fork of cot-transparency. We replace their remote training service \citep[Tinker;][]{tinker2025} with a local implementation of its interface. Around the trainer, we supply our own prompts and chat templates (for \llama{} the Hugging Face template, which adds a short system block that the authors' renderer omits). We add a bare-letter fallback to the answer parser, because theirs reads almost no bare-letter answers, and we run the epoch in six segments through the trainer's resume path. Every method trains a rank-8 LoRA adapter for one epoch.
\begin{tightlist}
\item \emph{BCT} adapts q, k, v and o ($\alpha{=}16$). Its loss is cross-entropy over prompt and response tokens, as in the GPT-3.5 runs of the BCT paper, so most loss tokens are question text. It uses batch 2 with 8-step accumulation.
\item \emph{ACT} adapts q and v (on \gemma{} inherited from the base configuration of the repository, whose Gemma-2-2B ACT configuration is labeled full fine-tuning). Its loss is the squared residual distance to the uncued activations of the frozen initial model, over the longest shared token suffix and all layers, with weight $10^{-4}$ and batch 1.
\item \emph{AttCT} adapts q and v. Its loss is the Jensen--Shannon divergence between per-head attention rows on the uncued content span, over all layers, with batch 1.
\item \emph{RMCT} adapts attention, MLP and unembedding weights (the default modules of the trainer), with 128 cued and 128 uncued rollouts per question. The reward is $r=-(\hat p-p_{\mathrm{ref}})(y-\hat p)$, where $y=1$ if a rollout gives the cued letter and $\hat p$ and $p_{\mathrm{ref}}$ are the cued and uncued rates, and advantages are standardized over the parsed cued rollouts of each batch. Unparsed rollouts drop out of the rates, the gradient and the KL term. The loss is the PPO-clipped loss of the trainer (GRPO in the terms of the paper), with KL coefficient 0.05, 16 generated tokens and 4 questions per batch. The trainer skips batches with zero advantage, so the 304 batches give 294 to 303 updates in the final RMCT runs (287 to 297 for RMCT+anchor). The LoRA scale is 4 ($\alpha{=}32$), as the Tinker cookbook documents. Tinker does not expose $\alpha$, and the RMCT paper states $\alpha{=}16$ on attention and MLP.
\end{tightlist}
The ACT and AttCT dataloaders truncate 5--7 prompts longer than 512 tokens. The exploratory RMCT+anchor variant adds the optional anchor reward of the trainer at its code default weight of 0.5, with 128 anchor rollouts, which the RMCT paper does not use. We declared it after every pre-registered \gemma{} RMCT learning rate evaluated by then had collapsed to a constant answer.

Our runs also depart from the source recipes in other ways. \citet{irpan2025consistency} fine-tune all self-attention parameters with batches of 4 or 8. The AttCT paper states q, k, v, o LoRA targets in its method section and 8-step accumulation in its ablations, unlike the repository configuration we run. The selected BCT learning rates are 100 and 10 times those of the BCT configurations in the repository (1e-6 for the Gemma-3-4B configuration we copy and 5e-6 for \llama). Library versions differ from the authors' lock file (Appendix~\ref{app:artifacts}).

\subsection{BCT variants}
All variants use the BCT pipeline. Bayes-BCT samples each target from $p_0(a\mid q)\,L(h\mid a)$, where $p_0(a\mid q)$ is the original model's uncued answer distribution and $L(h{=}X\mid a)=r$ if $a{=}X$ and $(1-r)/(K-1)$ otherwise, for $K$ options (an unparsed target is kept). Gold-gated BCT keeps the model's own targets when they are correct, corrects wrong targets when the cue is right and drops the rest. Position-diverse BCT places each training cue at a random one of the four positions. Response-only BCT masks the prompt from the loss and drops the instruction mix (\gemma{} only). The self-training control trains on the same targets and instruction mix with the cue removed. Response-only BCT has its own learning-rate sweep, and the other variants and the control use the rate selected for BCT on the same model.

\subsection{Learning-rate selection}
\label{app:selection}
\begin{table*}[t]\centering\scriptsize\setlength{\tabcolsep}{4pt}
\tabSelection
\caption{Stored and re-selected learning rates, with the validation F1 at every grid point (unmerged adapters, generated answers). $^d$ marks a degenerate point (excluded), and $^\dagger$ an exploratory variant.}
\label{tab:selection}
\end{table*}
We sweep learning rates on seed 42 and select them by the F1 score of \citet{irpan2025consistency}, the harmonic mean of uncued accuracy and the rate of resisting a wrong cue on validation items. The pre-registered rule has three parts. A best F1 at a grid end extends the grid by one half-decade, at most twice (once for response-only BCT, as its protocol specifies). A tie within 0.005 goes to the reference rate if that rate is among the tied points, and otherwise to the best F1. A checkpoint that gives one letter on at least 95\% of uncued validation prompts is degenerate and excluded. We apply this last rule to generated answers on \gemma, with unparsed answers as a category, and to the first-position readout on \llama.

The reference rates (called ``published'' in the prediction texts) are those of the authors' configurations. On \gemma, they are 1e-6 for BCT (the Gemma-3-4B configuration we copy) and for ACT (the repository's Gemma-2-2B configuration, chosen there by the same F1), and 5e-6 for AttCT (the default its configurations inherit). Every \llama{} method except RMCT also has 5e-6. For RMCT, the reference rate is 2.86e-4, the middle of the three rates its paper pools. \citet{irpan2025consistency} publish no rates.

We first selected rates on merged checkpoints (with the first-position readout on \gemma) and re-applied the same rule to the unmerged sweep after the merge correction (Table~\ref{tab:selection}). Eight of the ten selections are unchanged. \gemma{} ACT moves from 3.16e-6 to 1e-5 and AttCT from 5e-6 to 1.58e-5, and all reported results for these two methods use the re-selected rates. At the superseded rates, their unmerged test rates are \vGemActOldTrained{} (trained cue), \vGemActOldHeld{} (held-out) and \vGemActOldAcc{} (accuracy) for ACT, and \vGemAttctOldTrained, \vGemAttctOldHeld{} and \vGemAttctOldAcc{} for AttCT. The draft of the tie rule before Amendment 1 would have kept 3.16e-6 for \gemma{} ACT, which lies within the 0.005 tie band of 1e-5.

Several selections are weak. The F1 curves of ACT and AttCT vary by about .03 or less across their grids on both models, so these selections are weak preferences. The OV depth centroid of ACT is \vGemActOVOld{} at the superseded rate and \vGemActOV{} at the re-selected one (AttCT \vGemAttctOVOld{} and \vGemAttctOV). \gemma{} RMCT is selected at the grid edge after both permitted extensions. On \llama, RMCT at 2.86e-4 stops producing the answer format in our 16-token direct-answer setting (its parse rate is near zero), which the one-letter rule cannot detect (L5).

\subsection{Scoring and statistics}
Answers are greedy 16-token generations parsed to a letter, and an unparsed answer counts as no answer. The two few-shot cues are regenerated with 128 tokens and scored at the test slot (Appendix~\ref{app:fewshot}). We evaluate all adapters unmerged, as in training, except in the RMCT check in the paper's regime. An item is \emph{known} if the uncued answer is correct. Harmful-cue following (\flipk) is the share of known items whose cued answer is the cued letter. The held-out rate is the unweighted mean over the five held-out cue types, and we measure uncued accuracy on the trained-cue items.

Test-set intervals are percentile intervals of a paired bootstrap that resamples whole questions within each dataset (B = \vBootB, and 2000 in the exploratory tables). The original model and every method and seed use the same resample. Relapse fractions use seed $t$ intervals and a binomial bootstrap, and patch recovery uses Wilson intervals \citep{wilson1927probable}. Test-set contrasts between methods combine the paired item SE with a seed SE. An independent script re-derives every point estimate in the primary tables from the per-item files.

\appflush
\section{Instruments}
\label{app:instruments}
\subsection{Cue battery}
\begin{figure}[t]
\centering\footnotesize
\fbox{\parbox{0.94\columnwidth}{\raggedright\linespread{1.05}\selectfont
\cuepos{front} \textit{I believe the best solution is B, any disagreements?}\\
Find the product of $-8\cdot(-4)$.\\[1pt]
\cuepos{pre}\\
Answer choices: (A) $-4$\quad(B) $4$\quad(C) $32$\quad(D) $-32$\\[1pt]
\cuepos{post}\\
Answer with just the single letter (A, B, C, D, ...), no explanation or reasoning.\\[1pt]
\cuepos{end}}}
\caption{\textbf{The four cue positions in an example prompt} (a training question). Training places every cue at \emph{front}. Here a harmful cue names B and a helpful cue names the correct answer C (options shown on one line).}
\label{fig:prompt}
\end{figure}
The cue battery has \vGemGenNItems{} uncued suggested-answer questions (100 per dataset, rows 1200--1300, or the last 100 rows of shorter dumps), and it shares no questions with training. Each question gets six cue wordings (three from the training family and three novel) at the four positions of Figure~\ref{fig:prompt}, and each cue names a wrong option $X$ or the correct option $T$. We read the answer at the first position with the v7.1 readout, which Appendix~\ref{app:gates} defines and validates against generation. Helpful-cue following is P(answer $T$ $\mid$ wrong uncued answer, cue names $T$), and harmful-cue following is P(answer $X$ $\mid$ correct uncued answer, cue names $X$), both averaged over wordings. For the paired drops, helpful-cue following uses the items that both the original model and the checkpoint answer wrongly without a cue, and harmful-cue following uses the items that both answer correctly. The ratio, or content-awareness index (CAI), is the helpful drop divided by the harmful drop, computed where the harmful drop exceeds .02 (Table~\ref{tab:variants} gives it wherever the harmful drop is positive). A method is content-aware (selective) if its harmful drop is at least .05 and its ratio is at most .5. The relative reduction at a position is $1 -$ method/original harmful-cue following at that position, per seed.

\subsection{Circuit patching}
We fixed the circuit heads before any training run reported here. On \gemma{} they are the readers L18.H6, L14.H2, L13.H6 and L13.H2 and the writers L17.H3, L16.H4 and L15.H7, all in the layer band of the circuit (13--18). \citet{gupta2026many} name the first three readers and the writer hub L17.H3, and we added the other three from head-level patching and attention to the cued letter in the original model in that band. On \llama{} they are the writer hub L17.H24, L18.H28 (the first of the example readers named there) and L15.H8. In that work, L15.H8 and L17.H24 are the two top heads in every leave-one-cue-out fold. That work leaves the role of L15.H8 open. Our \llama{} subsets group it with L18.H28 as ``readers'' and leave L17.H24 alone as ``writers''. We patch the remaining cue-induced flips of each model, which are the items it answers correctly without the cue and with the cued letter when cued. In the cued run, we replace the answer-position activations of the heads with their values from the uncued run. Recovery is the share of flips whose answer returns to $T$. Three random head sets matched to the layers of the circuit (the same for every checkpoint) give the baseline.

\subsection{Layer swap}
The q, k, v and o projections of the trained model (adapter unmerged) replace those of the original model in one block of layers. The blocks are EARLY 0--6, CARRIER 7--12, CIRCUIT 13--18 and LATE 19--25 on \gemma{} and 0--8, 9--14, 15--18 and 19--31 on \llama. We use items that the original model answers correctly without the cue and with the cued letter when cued, and that the trained model answers correctly without the cue (at most 10 per cue and dataset). Let $h$ be harmful-cue following on these items. The forward mediated fraction $(1-h_{\mathrm{spliced}})/(1-h_{\mathrm{trained}})$ approximates the share of the trained model's fixes that the mixed model reproduces. The fixed items need not coincide, so Table~\ref{tab:route} also gives the item-level share. It is defined when the trained model fixes at least 20 items. Fractions for single blocks are not additive, because two blocks can each reproduce most of a fix. RMCT cannot be tested this way, because its adapter also trains the MLPs and the unembedding.

\subsection{Attention reading and weight-edit depth}
For every head, we take the attention mass from the final position to the cue span, which is the set of tokens inserted or changed in the cued prompt. The early readers of the \gemma{} circuit are \vGemEarlyReaders. The relative change is mass / original mass $-$ 1 on the labeled items of the original model, which are items it knows and on which it either followed the cue or strictly resisted it. For the weight-edit depth, we compute for each attention head the relative change of the OV and QK composites under the LoRA edit (attention projections only, which are q and v for ACT and AttCT). The depth centroid is the layer index weighted by the part of these changes above the median.

\subsection{Letter-prior control}
We fit a per-letter additive logit offset to the uncued log-probabilities of the original model, so that its shares of uncued answer letters approximately match those of the method. We score this shifted model with the same \flipk{} definitions, with the argmax restricted to the true options of each item. The drop beyond the prior is the control minus the method.

\subsection{Relapse fine-tuning}
Starting from each final checkpoint, with its CT adapter frozen, we train a fresh LoRA (rank 16, $\alpha{=}32$, on q, k, v, o, gate, up and down) with AdamW ($10^{-4}$), batch 16 and 125 steps. The data are 2,000 seeded examples from the AttCT instruction file, disjoint from the training mix of BCT, with the loss on assistant tokens. We re-run the test set with generated answers at step 125. At step 0 the first-position readout must reproduce the evaluation of the method exactly, and the generated answers are those of that evaluation (the original model is re-generated at step 0). The relapse fraction is (method at 125 $-$ method at 0) / (original at 0 $-$ method at 0), and the drift-adjusted fraction subtracts the change of the original model itself from the numerator.

\appflush
\section{Evaluation corrections}
\label{app:corrections}
We made three corrections during the study. We declared each in the record before computing the corrected results in full, with the exceptions listed in Appendix~\ref{app:prereg}. Every reported number uses the corrected procedures, with two exceptions. The held-out relapse columns and the merge comparison use the 16-token answers before the few-shot correction, and the RMCT check in the authors' regime scores merged weights.

\subsection{First-token readout and validity gates}
\label{app:gates}
Early instruments read the answer as the letter with the largest summed probability over its single-token forms (``L'' and `` L'') at the first generated position. \llama{} often answers ``(B)'', and its tokenizer has single tokens ``(L''. Ignoring these tokens, the readout disagreed with generation on 6.9\% (original), 19.7\% (BCT) and 11.6\% (ACT) of \llama{} test prompts, and an apparent rise of the held-out harmful-cue following of BCT on \llama{} was this artifact. Adding the ``(L'' forms was not enough for BCT, because the letter with the largest \emph{summed} probability is not always the letter of the most probable token, which is what greedy decoding writes. On \gemma, the correction moves the trained-cue and held-out rates of the original model from .500 and .602 to .490 and .596.

All test rates therefore use generated answers. The forward-pass instruments of \llama{} and the cue battery of both models use the maximum over letter-form logits with greedy tie-breaking (``v7.1''), and the other forward-pass instruments of \gemma{} keep the summed readout. A gate requires each checkpoint to agree with generation on at least .97 of the test prompts. All \vLlaGateN{} \llama{} checkpoints pass (agreement at least 0.99). The \vGenGateN{} cue-battery gates also pass (minimum over positions at least \vGenGateMin). We ran these on every 4th item with all cued prompts, for the original models, the seed-42 checkpoints of BCT, ACT, AttCT, RMCT and position-diverse BCT on both models, and response-only BCT on \gemma. We did not gate the cue-battery checkpoints of seeds 43 and 44 or of the remaining variants. Of the \vGemGateN{} \gemma{} checkpoints, \vGemGateNFails{} fail on the summed readout (unmerged, lowest agreement \vGemGateMin, \vGemGateFailList). Their letter-prior control and patch recovery use the v7.1 readout, against an original-model reference on the same readout. After benign fine-tuning the first-token readout also fails, so we score relapse on generated answers.

\subsection{Few-shot cues at the test slot}
\label{app:fewshot}
On the \emph{wrong few-shot} and \emph{spurious few-shot squares} cues, both models often answer every demonstration in turn, so a parser that reads the first line returns the answer to the first demonstration. We regenerated these prompts with 128 tokens (checking that the first 16 tokens equal the stored ones) and scored them at the test slot by a declared rule. This lowers the held-out mean of the original \gemma{} from \vFsGemBaseHeldStored{} to \vFsGemBaseHeldCorr{} and moves the held-out mean of every trained method by at most \vFsMaxArmShift. Counting a list output as no answer (U) or dropping items whose list has no test slot (CC) bounds the effect. The bounds are widest for the squares cue on \llama, where list outputs are common (original U \vFsLlaSqU, CC \vFsLlaSqCC). The held-out difference of \llama{} BCT from the original model is near zero under the reported rule and takes either sign across the bounds (U \vFsLlaBctHeldDiffU, CC \vFsLlaBctHeldDiffCC). The forward-pass instruments read the first position and cannot apply this rule, so we recomputed them without the two few-shot cues. Drops beyond the letter prior change by at most \vFsBpdMaxChange{} and patch recoveries by at most \vFsPatchMaxChange{} (\llama{} BCT from \vFsLlaBctPatchAll{} to \vFsLlaBctPatchNoFs, against \vFsLlaBasePatchAll{} to \vFsLlaBasePatchNoFs{} for the original model), and no conclusion in \S\ref{sec:mechanism} changes.

\subsection{Merged versus unmerged adapters}
\label{app:merge}
\begin{table*}[t]\centering\scriptsize\setlength{\tabcolsep}{4pt}
\tabMerge
\caption{Merged (bf16 \texttt{merge\_and\_unload}) and unmerged adapters, per method over its three seeds. The columns give the range of agreement of the generated test answers, the seed-mean difference (merged $-$ unmerged) in trained-cue \flipk, held-out \flipk{} and uncued accuracy, and the number of seeds that meet the declared rule. The merge check came before the learning-rate re-selection, so the \gemma{} ACT and AttCT rows are the checkpoints at the superseded rates. Held-out rates use the 16-token answers before the few-shot correction. $^\dagger$ marks an exploratory variant.}
\label{tab:merge}
\end{table*}
Early instruments merged each LoRA adapter into the bf16 base weights, while training and the authors' evaluator use it unmerged. Before inspecting any result, we declared that a merged result stands only if the generated test answers of the merged and unmerged checkpoints agree on at least .99 of the 3,406 test prompts \emph{and} each of the three headline rates differs by at most .02. Only \vMergeNStand{} of \vMergeN{} checkpoints meet this rule (agreement \vMergeAgreeMin--\vMergeAgreeMax, Table~\ref{tab:merge}). Most rate differences are small, but merging weakens RMCT on both models (its merged trained-cue and held-out rates are higher), consistent with its small, spread-out update being rounded away in bf16. On merged weights, the \llama{} layer swap gives the same D1 verdicts (LATE block BCT \vLlaBctRouteMergedLate, ACT \vLlaActRouteMergedLate, AttCT \vLlaAttctRouteMergedLate). Every result therefore uses unmerged adapters, except the RMCT check in the paper's regime (Table~\ref{tab:rmctpaper}), which the authors' vLLM pipeline evaluates on merged weights. Because merging weakened RMCT in our setting, its effect there may be understated.

\appflush
\section{Additional results}
\label{app:extra}
This appendix gives the tables behind Sections~\ref{sec:efficacy}--\ref{sec:relapse}, in the same order. Analyses outside the pre-registration are marked exploratory.

\subsection{Efficacy and transfer (\S\ref{sec:efficacy})}
\label{app:percue}
\begin{table*}[t]\centering\scriptsize\setlength{\tabcolsep}{3pt}
\tabMain
\caption{\textbf{All four methods reduce following of the trained cue, and transfer depends on the method and the model.} Test set, generated answers. Harmful-cue following (flip$\mid$\knew) for the trained cue at its original positions and averaged over the five held-out cue types, and uncued accuracy. Seed mean $\pm$ SD. $\Delta$ is the method minus the original model, with its 95\% question-cluster interval (no interval for accuracy). The cells of the original model show their own interval.}
\label{tab:main}
\end{table*}
\begin{table*}[t]\centering\scriptsize\setlength{\tabcolsep}{4pt}
\tabPerCue
\caption{Per-cue \flipk{} on the test set (generated answers, with few-shot cues scored at the test slot). The table gives the original value and, for each method, the seed mean with the change from the original model in parentheses. $^{\circ}$ marks a change whose 95\% question-cluster bootstrap interval includes zero, and $^\dagger$ an exploratory variant.}
\label{tab:percue}
\end{table*}
\begin{table*}[t]\centering\scriptsize\setlength{\tabcolsep}{4pt}
\tabLetterPrior
\caption{Letter-prior control (exploratory, first-position readout, with the gate-failing \gemma{} checkpoints on the v7.1 readout). The table gives the \flipk{} of the method and of the original model shifted to the uncued letter distribution of the method, on the trained cue and the held-out mean. It also gives the drop beyond the prior (control $-$ method, seed mean $\pm$ SD) and the share of uncued answers on the most frequent letter of the method (seed mean, with the letter per seed). $^\dagger$ marks an exploratory variant.}
\label{tab:letterprior}
\end{table*}
Table~\ref{tab:main} gives the test-set rates behind Figure~\ref{fig:efficacy}. The held-out mean hides opposite movements on individual cues (Table~\ref{tab:percue}). BCT \emph{raises} harmful-cue following on the distractor-argument cue on both models while lowering most of the others. Only RMCT and its anchored variant significantly lower it on the wrong few-shot cue on either model, counting an unparsed answer as not following the cue. In the letter-prior control (Table~\ref{tab:letterprior}), the drop beyond the letter prior is positive on the trained cue for every method, and on held-out cues it is near zero only for \llama{} BCT. \gemma{} RMCT puts about half of its uncued answers on one letter (D in every seed), against at most about a third for every other method, so we read its raw rates next to this control.

\appflush
\subsection{Selectivity (\S\ref{sec:content})}
\label{app:selectivity}
\begin{table*}[t]\centering\scriptsize\setlength{\tabcolsep}{3pt}
\tabContentPosition
\caption{\textbf{Helpful cues drop together with harmful ones, and the fix fades when the cue moves after the options.} Cue battery, pooled over six wordings and four positions. Paired drops in helpful- and harmful-cue following (seed mean $\pm$ SD), their ratio with a 95\% question-bootstrap interval (``--'' where the harmful drop exceeds .02 in fewer than 97.5\% of samples), and the relative reduction of harmful-cue following at each position (only front is trained).}
\label{tab:content}
\end{table*}
\begin{table*}[t]\centering\scriptsize\setlength{\tabcolsep}{4pt}
\tabScale
\caption{Selectivity on three scales. Each cell is the seed-mean helpful drop divided by the seed-mean harmful drop on the paired item sets, over all positions, with a 95\% question-bootstrap interval (B = 2000, and ``--'' where the harmful drop exceeds .02 in fewer than 97.5\% of samples). Exploratory. $^\dagger$ marks an exploratory variant.}
\label{tab:scale}
\end{table*}
\begin{table*}[t]\centering\scriptsize\setlength{\tabcolsep}{4pt}
\tabContentByPos
\caption{Helpful-to-harmful ratio before the options (front and pre pooled) and at each position, with each cell on three scales (``--'' on a scale where the harmful drop is below .02). Seed means on the paired item sets. Exploratory. $^\dagger$ marks an exploratory variant.}
\label{tab:contentbypos}
\end{table*}
\begin{table*}[t]\centering\scriptsize\setlength{\tabcolsep}{4pt}
\tabSameItem
\caption{Same questions, cue before the options. The table gives the following of a cue that names the correct answer and of a cue that names a wrong option on the same questions, original $\to$ trained (seed means). The questions are those that both the original and the trained model answer wrongly without a cue, and the option that the wrong cue names is neither model's uncued answer. The last columns give the difference of the two drops and the change in their log-odds gap, with 95\% question-bootstrap intervals. Exploratory.}
\label{tab:sameitem}
\end{table*}
Table~\ref{tab:content} gives the paired drops and their ratio pooled over positions. Table~\ref{tab:scale} recomputes the ratio on the same item sets as absolute drops, proportional drops ($1-$ trained/original rate) and log-odds shifts. The original model follows correct cues more often than wrong ones, so the absolute scale favors a larger helpful drop and the proportional scale a smaller one. Pooled over positions, every method stays far above the ratio of 0.5 that would mark a selective fix on every scale. Table~\ref{tab:contentbypos} gives the ratios position by position, and Table~\ref{tab:sameitem} gives the same-question test. All of these analyses are exploratory.

Two exceptions appear. After the options, where the fix is weak, the helpful drop is mostly the smaller one in absolute and proportional terms, and at the end position a few ratios fall below one half (Table~\ref{tab:contentbypos}). The second exception is in the same-question test, with the cue before the options. There, on the log-odds scale, the little cue following that remains favors correct cues more than before for \gemma{} RMCT and for \gemma{} gold-gated BCT, an oracle variant (Table~\ref{tab:sameitem}).

\appflush
\subsection{Position (\S\ref{sec:position})}
\label{app:position}
\begin{table*}[t]\centering\scriptsize\setlength{\tabcolsep}{4pt}
\tabPosCheck
\caption{Position checks (seed means, exploratory). \emph{Matched} is the relative reduction of harmful-cue following in wording $\times$ position cells whose original harmful-cue following lies in the range shared by cells before and after the options. \emph{After coef.} is the after-the-options coefficient of a regression of relative reduction on original harmful-cue following and position over the 24 cells. \emph{End, non-correct} is the end-position relative reduction when every answer other than the correct one counts as following the cue. The right part gives the relative reduction by position for the three training-family wordings and the three novel wordings. $^\dagger$ marks an exploratory variant.}
\label{tab:poscheck}
\end{table*}
\begin{table*}[t]\centering\scriptsize\setlength{\tabcolsep}{4pt}
\tabLayout
\caption{The trained cue of the test set (Table~\ref{tab:main}) split by the cue position of each source row (post hoc, exploratory). The table gives the original \flipk{} per position with the number of known items in parentheses, the relative reduction of \flipk{} per position (seed means), and the front$-$post gap with a 95\% question-bootstrap interval (B = 2000, with the same resampled questions for the original model and every seed). $^\dagger$ marks an exploratory variant.}
\label{tab:layout}
\end{table*}
\begin{table*}[t]\centering\scriptsize\setlength{\tabcolsep}{3pt}
\tabLRSweep
\caption{Robustness of the pattern to the learning rate and the training data (exploratory, seed 42). The table gives the paired drops on the cue battery, the helpful-to-harmful ratio on three scales and the relative reduction of harmful-cue following by position. It covers the selected learning rate, its non-degenerate neighbors, and the \llama{} checkpoints trained with the authors' own data and configuration.}
\label{tab:lrsweep}
\end{table*}

\paragraph{Other explanations.} Table~\ref{tab:poscheck} tests three other explanations of the position effect. \emph{Cue strength.} Cues after the options are on average stronger. We therefore select the wording $\times$ position cells whose original harmful-cue following lies in a range shared by cells before and after the options (\gemma{} \vGemMatchLo--\vGemMatchHi, with only \vGemMatchNAfter{} cells after the options, and \llama{} \vLlaMatchLo--\vLlaMatchHi). In these cells, the reduction is still much larger before the options than after them for BCT, ACT and AttCT, and larger by less for RMCT. A regression of the reduction on strength and an after-the-options indicator over all 24 cells agrees. \emph{A third letter.} If we count every answer other than the correct one as following the cue, the reduction of \gemma{} RMCT falls from \vGemRmctRelredPost{} to \vGemRmctNCPost{} just after the options and from \vGemRmctRelredEnd{} to \vGemRmctNCEnd{} at the end, and that of the other methods falls less. If we also leave out the questions whose correct answer is each checkpoint's most frequent uncued letter, \gemma{} RMCT removes \vGemRmctNCxPost{} and \vGemRmctNCxEnd{} (BCT \vGemBctNCxPost{} and \vGemBctNCxEnd). Under the same count, \llama{} RMCT keeps \vLlaRmctNCxPost{} and \vLlaRmctNCxEnd{} (ACT \vLlaActNCxPost{} and \vLlaActNCxEnd). Across the three \gemma{} seeds of RMCT, its share of uncued \vGemRmctModalLetter{} answers on the cue battery (\vGemRmctModalShareSeeds) and its unadjusted reduction just after the options (\vGemRmctPostSeeds) fall together. The anchored variant, which has no strong letter preference, keeps \vGemRmctaNCxPost{} and \vGemRmctaNCxEnd{} on \gemma, about as much as BCT just after the options and more at the end. \emph{Wording.} Before the options, the three novel wordings, which never occur in training, are reduced within about .06 of the three training-family wordings (\llama{} BCT reduces the novel ones more), and after the options both families fade together. The training family has 51 templates, which limits this contrast, but it suggests that the position of the cue matters more than its phrasing.

\paragraph{Learning rate and training data.} Table~\ref{tab:lrsweep} runs the cue battery on the seed-42 checkpoints one grid step below and above the selected learning rate of each of the four methods (where that point exists and is not degenerate) and on the \llama{} BCT, ACT and AttCT checkpoints trained with the data and configuration of the AttCT authors (\vSweepN{} checkpoints, declared before the runs in Table~\ref{tab:amend}). At every learning rate, the helpful drop is at least as large as the harmful drop in absolute terms and at least \vSweepRatioMin{} times it on every scale, and for BCT, ACT and AttCT the fix stays tied to the positions before the options. ACT and AttCT trained with the authors' recipe show the same pattern, while BCT trained with the authors' recipe reduces harmful-cue following by at most about a third at any position.

\paragraph{Original cue positions.} Table~\ref{tab:layout} splits the trained-cue items of the test set by the cue position that each source row already has, an analysis we ran after all other results were known. Of these items, \vLayNFront{} put the cue before the question (the trained position), \vLayNPre{} before the options and \vLayNPost{} after them. With the cue first, every method removes \vLayFrontMin--\vLayFrontMax{} of harmful-cue following. After the options, ACT and AttCT remove only \vGemAttctLayPost--\vLlaActLayPost{} on both models, as on the cue battery. RMCT keeps most of its fix there on both models (without a letter control on \gemma), and BCT keeps it on \gemma{} only. The interval of the front$-$post gap excludes zero except for \gemma{} BCT and \llama{} RMCT, and the cells are small (30--43 known items per position and model).

\appflush
\subsection{Mechanism (\S\ref{sec:mechanism})}
\label{app:mechanism}
\begin{table*}[t]\centering\scriptsize\setlength{\tabcolsep}{4pt}
\tabPatch
\caption{Patch recovery. Each cell is the share of a method's remaining cue-induced flips that return to the correct answer when the uncued answer-position activations of a head set are patched in, pooled over seeds (\gemma{} three seeds, \llama{} seed 42). The head sets are the circuit (with a Wilson 95\% interval), its writer and reader subsets (the \llama{} writer subset is L17.H24 alone), the circuit on trained-cue flips only, and the best random head set (the maximum over three layer-matched sets and, for \gemma, over seeds). The \gemma{} circuit column and the subset and trained-cue columns are exploratory.}
\label{tab:patch}
\end{table*}
\begin{table*}[t]\centering\scriptsize\setlength{\tabcolsep}{3pt}
\tabReading
\caption{Early reader heads of \gemma. The left part gives the relative change of their cue-span attention on the labeled items of the original model (seed mean $\pm$ SD, with held-out cues in the order post hoc, distractor fact, wrong few-shot, distractor argument and squares). The middle part (exploratory) uses the trained-cue items that the original model follows. It gives the relative drop on the items the method fixes (its cued answer is no longer the cued letter) and on the items it still follows (seed means), the fixed-item drop as a share of the gap of the original model between items it follows and items it resists, and the item counts per seed. The right part gives the OV depth centroid of the LoRA edit over the attention projections (layer index out of 26). $^\dagger$ marks an exploratory variant.}
\label{tab:reading}
\end{table*}
\begin{table*}[t]\centering\scriptsize\setlength{\tabcolsep}{4pt}
\tabRoute
\caption{Layer swap. Each cell is the forward mediated fraction when one block of the trained model's q, k, v and o projections is copied into the original model (\llama{} EARLY 0--8, CARRIER 9--14, CIRCUIT 15--18 and LATE 19--31, \gemma{} 0--6, 7--12, 13--18 and 19--25), on items whose cue the original model follows. The three readouts are the pre-registered binary readout, the share of the trained model's own fixes and the share of the fall in the probability of the cued letter. For \llama{} (seed 42), the table also gives D1 (the largest block is EARLY or CARRIER, and LATE is below 0.15). For \gemma, it gives seed means over three seeds, the range over seeds of the number of fixed items, and the number of seeds that meet the same criterion (exploratory). RMCT cannot be tested this way.}
\label{tab:route}
\end{table*}
Table~\ref{tab:patch} gives patch recovery for the circuit, its head subsets and the trained-cue flips alone. Table~\ref{tab:reading} gives the cue-span attention of the early readers on the labeled items of the original model (the quantity that P3 tests), its drop on the trained-cue items that each method fixes and on those it still follows (exploratory), and the depth centroids that P4 tests. For ACT and AttCT, the gap between fixed and still-followed items appears only on the trained cue. Pooled over all six cue types, the attention drops on the two kinds of items are about equal for every method (exploratory). The OV depth centroids of \gemma{} order AttCT $<$ ACT $<$ RMCT in every seed, so P4 holds. However, the centroid of ACT moves between the two learning rates selected for it (Appendix~\ref{app:selection}), and the methods place their adapters on different modules, so we do not treat depth as a property of the objective. Table~\ref{tab:route} gives the layer swap on both models under three readouts. These are the pre-registered binary readout, the share of the trained model's own fixes that the mixed model reproduces, and the share of the fall in the probability of the cued letter. The \gemma{} layer swap is exploratory, because D1 concerns \llama.

\appflush
\subsection{Training-data variants (\S\ref{sec:variants})}
\label{app:variants}
\begin{table*}[t]\centering\scriptsize\setlength{\tabcolsep}{3pt}
\tabVariants
\caption{\textbf{Changing the training data of BCT.} Seed means for the BCT variants of Section~\ref{sec:variants} (defined in Appendix~\ref{app:details}). Response-only BCT is run on \gemma{} only. CAI (content-awareness index) is the helpful drop divided by the harmful drop ($^{*}$ marks a harmful drop below .05, which the registered criterion does not count as selective). Rel.\ red.\ is the relative reduction of harmful-cue following at the front and end positions.}
\label{tab:variants}
\end{table*}
Table~\ref{tab:variants} gives the values behind Figure~\ref{fig:variants} for both models, with test-set rates and uncued accuracy.

\appflush
\subsection{Relapse (\S\ref{sec:relapse})}
\label{app:relapse}
\begin{table*}[t]\centering\scriptsize\setlength{\tabcolsep}{2.5pt}
\tabRelapse
\caption{Relapse under 125 steps of benign LoRA fine-tuning (generated answers, three seeds per method, $\pm$ is the seed SD). The relapse fraction is (method at step 125 $-$ method at step 0)/(original at step 0 $-$ method at step 0), and its seed 95\% CI is a $t$ interval over seeds. \emph{Beyond original drift} subtracts the change of the original model under the same fine-tuning. \emph{Log-odds rise} is the change in the logit of trained-cue \flipk{} from step 0 to 125, with its ratio to that of the original model in parentheses. The drift-adjusted and log-odds columns are exploratory. Held-out fractions are given only where the held-out gap of the method is at least .05. The held-out columns use the 16-token answers without the few-shot correction (as does the original held-out rate in the header), and the held-out fraction beyond drift subtracts the held-out change of the original model.}
\label{tab:relapse}
\end{table*}
Table~\ref{tab:relapse} gives the relapse fractions behind \S\ref{sec:relapse} on the probability and log-odds scales. Beyond the drift of the original model, the fraction of \llama{} ACT has the seed 95\% interval \vLlaActRelAdjSeedCI{} and the item interval \vLlaActRelAdjItemCI. On the log-odds scale, RMCT, which starts from the lowest rate, rises two to three times as much as the original model.

\appflush
\section{Checks against the source papers}
\label{app:authors}
\begin{table*}[t]\centering\scriptsize\setlength{\tabcolsep}{3pt}
\tabAuthorsEval
\caption{The evaluator of the AttCT authors on seed-42 final checkpoints (adapter unmerged, their code). The columns give the MMLU bias rate (BRR = P(cued answer $\mid$ cued) $-$ P(cued answer $\mid$ clean), n = 1000) with its unpaired SE, the held-out BRR (n = 951), clean MMLU accuracy (Wilson 95\%) and the Anthropic model-written sycophancy rate (n = 999). \gemma{} ACT and AttCT use the re-selected learning rates (Appendix~\ref{app:selection}), and the \gemma{} BCT sweep rows are three other learning rates of the BCT grid. $^\S$ marks the authors' own data and configuration run through their pipeline. Values in parentheses are those that the AttCT paper reports for its Llama checkpoints.}
\label{tab:authors}
\end{table*}
\begin{table*}[t]\centering\scriptsize\setlength{\tabcolsep}{4pt}
\tabRmctPaper
\caption{RMCT in the authors' own regime (their trainer and paper configuration, chain-of-thought answers, evaluated with their protocol on 100 questions per pass). The table gives the switch rate toward the bias on the training bias and pooled over five held-out biases, for final-step checkpoints pooled over two evaluation passes. Next to these are the values recomputed from the evaluation logs that the authors released (one training run, two passes). The authors' base value comes from their 400-question logs, and their held-out value for the second pass pools four biases, while ours pools five. A second training run at 2.86e-4 (seed 43) gives \vRPSeedFtTrain{} and \vRPSeedFtHeld{} at its final step. We evaluate with the authors' vLLM pipeline on bf16-merged weights, which weakened RMCT in our setting (Appendix~\ref{app:merge}). The direction reproduces, but the size does not at 2.86e-4.}
\label{tab:rmctpaper}
\end{table*}
Two checks tie our pipeline to the source papers. First, on the evaluator of the AttCT authors (Table~\ref{tab:authors}), the \llama{} checkpoints trained with the authors' own data and configuration reproduce the values that the AttCT paper reports (its Table 11) within sampling error. \llama{} ACT and AttCT trained on our shared data come close to these values on MMLU BRR (\vAELlamaOursActSFourTwoBrr{} and \vAELlamaOursAttctSFourTwoBrr{} against \vAELlamaReproActBrr{} and \vAELlamaReproAttctBrr), but not on held-out BRR (\vAELlamaOursActSFourTwoHeld{} and \vAELlamaOursAttctSFourTwoHeld{} against \vAELlamaReproActHeld{} and \vAELlamaReproAttctHeld). According to the authors' evaluator code, that held-out rate uses held-out questions with the cue templates of their training data, which our shared data do not contain verbatim. It does not use held-out cue types, so the near-zero values of the AttCT paper and our partial transfer do not conflict. The clean accuracy of \gemma{} BCT falls from \vAEGemmaBaseAcc{} to \vAEGemmaOursBctSFourTwoAcc. BCT loses accuracy at every learning rate in the table that moves the MMLU bias rate (only nominally at 3.16e-4), whereas the selected full-attention BCT of \citet{irpan2025consistency} keeps its MMLU accuracy on Gemma~2 2B (61.6 against 61.0). Second, RMCT in the authors' own regime (Table~\ref{tab:rmctpaper}) reproduces the direction of their reported effect but not its size. At their middle learning rate, our two training seeds reach switch rates of \vRPLrBTrain{} (two evaluation passes) and \vRPSeedFtTrain{} (one pass) on the training bias, against \vRPPaperLrBDA{} in the authors' two evaluations of their run. This regime uses their prompts, renderer and parser. Possible causes of the gap are the variance of a 16-step run, scoring merged weights, and a difference between Tinker's internal LoRA step size, which we cannot inspect, and that of our local implementation.

\section{Artifacts, licenses and compute}
\label{app:artifacts}
\paragraph{Models.} We use \gemmafull{} (Gemma Terms of Use) and \llamafull{} (Llama 3.1 Community License) unmodified as base models for LoRA training, within their license terms and use policies. Any released adapters will carry the required notices.
\paragraph{Data.} Questions come from the multiple-choice dumps released with \citet{chua2024bct} (cot-transparency, MIT license), built on MMLU (MIT), HellaSwag (MIT per its dataset card), LogiQA (its paper states that it is freely available, and its repository has no license file) and TruthfulQA (Apache-2.0). The evaluator of the AttCT authors also uses the model-written sycophancy evaluations of Anthropic \citep[CC BY 4.0;][]{perez2022discovering}, and the RMCT check in the paper's regime uses the multiple-choice questions of Humanity's Last Exam \citep[MIT;][]{phan2025hle}, as the authors' pipeline does. These are English benchmark questions without personal information. The benign fine-tuning examples are 2,000 rows of the instruction file distributed with the AttCT repository, disjoint from the rows in the training mix of BCT. That file pairs instruction prompts with model-written responses, and the repository labels it ``Alpaca/GPT-4T'' without stating a license.
\paragraph{Code.} BCT, ACT and AttCT run from the public AttCT repository (no license file) at a fixed commit. RMCT runs the authors' trainer from their public fork of cot-transparency (MIT). We use the AttCT code unmodified and do not redistribute it, and our configurations are given as changes to its files. Our code, configurations and the redacted pre-registration file are in the supplementary material, and per-item results will be released for research use.
\paragraph{Compute and software.} All training and evaluation ran on single NVIDIA A100 GPUs (80\,GB, a few jobs on 40\,GB), for about 210 GPU-hours, including the retired pilot design. We used PyTorch 2.9.0, Transformers 4.57.6, PEFT 0.19.1, vLLM 0.11.2, Tinker 0.19.0 and tinker-cookbook 0.4.1.

\end{document}

%% file: numbers.tex
\newcommand{\missingnum}{\textbf{??}}

\newcommand{\vBasisNote}{UNMERGED adapters (primary results)}
\newcommand{\vIsMerged}{0}
\newcommand{\vBootB}{10000}

\newcommand{\vGemNItems}{1703}
\newcommand{\vGemNQuestions}{465}

\newcommand{\vGemGenNItems}{400}

\newcommand{\vGemAttctRatio}{1.32}

\newcommand{\vGemRmctRatio}{1.10}

\newcommand{\vGemRmctRelredPost}{0.542}

\newcommand{\vGemRmctRelredEnd}{0.382}

\newcommand{\vGemRmctPatch}{0.910}

\newcommand{\vGemRmctPatchRandom}{0.121}

\newcommand{\vGemRmctModalShare}{0.54}

\newcommand{\vGemEarlyReaders}{L13.H6, L13.H2, L14.H2}

\newcommand{\vGemActOV}{16.9}

\newcommand{\vGemAttctOV}{4.9}

\newcommand{\vGemBctCAI}{1.26}

\newcommand{\vGemBctroCAI}{1.15}

\newcommand{\vGemBayesEightCAI}{1.45}

\newcommand{\vGemGoldCAI}{1.06}

\newcommand{\vLlaBctRatio}{3.55}

\newcommand{\vLlaAttctRatio}{1.36}

\newcommand{\vLlaRmctRatio}{1.24}

\newcommand{\vLlaRmctFrontMinusPost}{+0.227}
\newcommand{\vLlaRmctFrontMinusPostCI}{\mbox{[0.193, 0.260]}}

\newcommand{\vLlaBctPatch}{0.598}

\newcommand{\vLlaBctRouteMergedLate}{0.25}

\newcommand{\vLlaActRouteMergedLate}{0.05}

\newcommand{\vLlaAttctRouteMergedLate}{0.09}

\newcommand{\vVerdictPOne}{FALSE}
\newcommand{\vVerdictPTwo}{FALSE}
\newcommand{\vVerdictPThree}{FALSE}
\newcommand{\vVerdictPFour}{TRUE}
\newcommand{\vVerdictPFive}{TRUE}
\newcommand{\vVerdictLOne}{TRUE}
\newcommand{\vVerdictLTwo}{TRUE}
\newcommand{\vVerdictLThree}{FALSE}
\newcommand{\vVerdictLFour}{TRUE}
\newcommand{\vVerdictLFive}{FALSE}
\newcommand{\vVerdictLFOne}{TRUE}
\newcommand{\vVerdictLFTwo}{TRUE}
\newcommand{\vVerdictLFThree}{TRUE}
\newcommand{\vVerdictLROne}{TRUE}
\newcommand{\vVerdictLRTwo}{FALSE}
\newcommand{\vVerdictDOne}{FALSE}
\newcommand{\vVerdictBOne}{TRUE}
\newcommand{\vVerdictBTwo}{FALSE}
\newcommand{\vVerdictBThree}{FALSE}
\newcommand{\vVerdictFOne}{FALSE}
\newcommand{\vVerdictFTwo}{FALSE}
\newcommand{\vVerdictFThree}{TRUE}
\newcommand{\vVerdictROne}{FALSE}
\newcommand{\vVerdictSOneGem}{FALSE}
\newcommand{\vVerdictSTwoGem}{TRUE}
\newcommand{\vVerdictSOneLla}{TRUE}
\newcommand{\vVerdictSTwoLla}{TRUE}

\newcommand{\vAEGemmaBaseAcc}{0.596}

\newcommand{\vAEGemmaOursBctSFourTwoAcc}{0.512}

\newcommand{\vAELlamaOursActSFourTwoBrr}{0.033}

\newcommand{\vAELlamaOursActSFourTwoHeld}{0.057}

\newcommand{\vAELlamaOursAttctSFourTwoBrr}{0.031}

\newcommand{\vAELlamaOursAttctSFourTwoHeld}{0.062}

\newcommand{\vAELlamaReproActBrr}{0.019}

\newcommand{\vAELlamaReproActHeld}{0.003}

\newcommand{\vAELlamaReproAttctBrr}{0.014}

\newcommand{\vAELlamaReproAttctHeld}{0.005}

\newcommand{\vRPLrBTrain}{0.362}

\newcommand{\vRPPaperLrBDA}{0.141/0.132}

\newcommand{\vRPSeedFtTrain}{0.350}
\newcommand{\vRPSeedFtHeld}{0.309}
\newcommand{\vGemGateN}{33}

\newcommand{\vGemGateNFails}{8}
\newcommand{\vGemGateMin}{0.956}
\newcommand{\vLlaGateN}{24}

\newcommand{\vGemGateFailList}{BCT (seed 44); response-only BCT (seeds 42, 43); Bayes-BCT with $r{=}.5$ (seed 44); Bayes-BCT with $r{=}.8$ (seed 44); gold-gated BCT (seeds 43, 44); position-diverse BCT (seed 43)}
\newcommand{\vGenGateN}{13}

\newcommand{\vGenGateMin}{0.990}

\newcommand{\vGemBctRelFracSeedCI}{\mbox{[$-$0.020, 0.394]}}

\newcommand{\vLlaActRelAdjItemCI}{\mbox{[$-$0.432, 0.513]}}
\newcommand{\vLlaActRelAdjSeedCI}{\mbox{[$-$0.021, 0.346]}}

\newcommand{\vMergeN}{51}
\newcommand{\vMergeAgreeMin}{0.924}
\newcommand{\vMergeAgreeMax}{0.991}
\newcommand{\vMergeNStand}{3}
\newcommand{\vMergeNFail}{48}
\newcommand{\vFsGemBaseHeldStored}{0.596}
\newcommand{\vFsGemBaseHeldCorr}{0.546}
\newcommand{\vFsMaxArmShift}{0.034}
\newcommand{\vFsLlaSqU}{0.065}
\newcommand{\vFsLlaSqCC}{0.539}
\newcommand{\vFsLlaBctHeldDiffU}{+0.044}
\newcommand{\vFsLlaBctHeldDiffCC}{\mbox{$-$0.068}}
\newcommand{\vFsBpdMaxChange}{0.07}
\newcommand{\vFsPatchMaxChange}{0.07}
\newcommand{\vFsLlaBctPatchAll}{0.598}
\newcommand{\vFsLlaBctPatchNoFs}{0.669}
\newcommand{\vFsLlaBasePatchAll}{0.650}
\newcommand{\vFsLlaBasePatchNoFs}{0.628}

\newcommand{\vGemMatchLo}{0.50}
\newcommand{\vGemMatchHi}{0.63}

\newcommand{\vGemMatchNAfter}{2}

\newcommand{\vLlaActMatchBefore}{0.82}
\newcommand{\vLlaActMatchAfter}{0.17}

\newcommand{\vLlaMatchLo}{0.26}
\newcommand{\vLlaMatchHi}{0.80}

\newcommand{\vLlaBeforeLogMin}{0.82}

\newcommand{\vSameDiffAllMin}{\mbox{$-$0.02}}
\newcommand{\vSameDiffAllMax}{+0.07}

\newcommand{\vGemBctNCxPost}{0.36}
\newcommand{\vGemBctNCxEnd}{0.14}

\newcommand{\vGemRmctNCPost}{0.39}
\newcommand{\vGemRmctNCEnd}{0.28}
\newcommand{\vGemRmctNCxPost}{0.18}
\newcommand{\vGemRmctNCxEnd}{0.10}

\newcommand{\vGemRmctaNCxPost}{0.38}
\newcommand{\vGemRmctaNCxEnd}{0.30}

\newcommand{\vLlaActNCxPost}{0.09}
\newcommand{\vLlaActNCxEnd}{0.05}

\newcommand{\vLlaRmctNCxPost}{0.63}
\newcommand{\vLlaRmctNCxEnd}{0.51}
\newcommand{\vGemRmctModalLetter}{D}
\newcommand{\vGemRmctModalShareSeeds}{0.615, 0.532 and 0.415}
\newcommand{\vGemRmctPostSeeds}{0.651, 0.542 and 0.433}

\newcommand{\vGemActReadStill}{2\%}

\newcommand{\vTrainCorrectCue}{239}
\newcommand{\vGemBctContra}{144}
\newcommand{\vLlaBctContra}{114}
\newcommand{\vGemActOVOld}{10.0}
\newcommand{\vGemAttctOVOld}{5.6}
\newcommand{\vSweepN}{18}
\newcommand{\vSweepRatioMin}{0.74}

\newcommand{\vLayNTotal}{216}
\newcommand{\vLayNFront}{68}
\newcommand{\vLayNPre}{68}
\newcommand{\vLayNPost}{80}

\newcommand{\vGemAttctLayPost}{0.14}

\newcommand{\vLlaActLayPost}{0.28}

\newcommand{\vLayFrontMin}{0.80}
\newcommand{\vLayFrontMax}{1.00}

\newcommand{\vGemActOldTrained}{0.260}
\newcommand{\vGemActOldHeld}{0.451}
\newcommand{\vGemActOldAcc}{0.457}

\newcommand{\vGemAttctOldTrained}{0.290}
\newcommand{\vGemAttctOldHeld}{0.485}
\newcommand{\vGemAttctOldAcc}{0.448}
\newcommand{\vRouteBelowMin}{49\%}
\newcommand{\vRouteBelowMax}{73\%}
\newcommand{\vRouteAboveMax}{8\%}

\newcommand{\vGemBctRelFracU}{0.187}
\newcommand{\vGemAttctRelFracU}{0.452}

\newcommand{\vLlaActRelAdjU}{0.162}

%% file: tables.tex
\newcommand{\tabMain}{%
\begin{tabular}{lrrrrr}
\toprule
method & trained-cue flip$\mid$knew & $\Delta$ [95\% CI] & held-out (5 cues) & $\Delta$ [95\% CI] & uncued accuracy ($\Delta$) \\
\midrule
\multicolumn{6}{l}{\textit{Gemma-2-2B-it}} \\
original & 0.490 [0.39, 0.59] & -- & 0.546 [0.50, 0.59] & -- & 0.454 [0.39, 0.52] \\
BCT & 0.116 $\pm$ 0.024 & -0.37 [-0.48, -0.27] & 0.421 $\pm$ 0.018 & -0.12 [-0.17, -0.08] & 0.406 $\pm$ 0.045 (-0.05) \\
ACT & 0.241 $\pm$ 0.012 & -0.25 [-0.34, -0.16] & 0.479 $\pm$ 0.004 & -0.07 [-0.10, -0.04] & 0.468 $\pm$ 0.005 (+0.01) \\
AttCT & 0.257 $\pm$ 0.010 & -0.23 [-0.32, -0.15] & 0.486 $\pm$ 0.003 & -0.06 [-0.08, -0.03] & 0.444 $\pm$ 0.005 (-0.01) \\
RMCT & 0.091 $\pm$ 0.015 & -0.40 [-0.50, -0.30] & 0.282 $\pm$ 0.043 & -0.26 [-0.31, -0.22] & 0.458 $\pm$ 0.028 (+0.00) \\
\midrule
\multicolumn{6}{l}{\textit{Llama-3.1-8B-Instruct}} \\
original & 0.555 [0.46, 0.64] & -- & 0.421 [0.39, 0.46] & -- & 0.551 [0.49, 0.62] \\
BCT & 0.350 $\pm$ 0.050 & -0.20 [-0.29, -0.12] & 0.421 $\pm$ 0.018 & +0.00 [-0.04, 0.04] & 0.532 $\pm$ 0.009 (-0.02) \\
ACT & 0.236 $\pm$ 0.029 & -0.32 [-0.41, -0.23] & 0.326 $\pm$ 0.007 & -0.10 [-0.12, -0.07] & 0.548 $\pm$ 0.005 (+0.00) \\
AttCT & 0.274 $\pm$ 0.014 & -0.28 [-0.37, -0.19] & 0.348 $\pm$ 0.003 & -0.07 [-0.10, -0.05] & 0.540 $\pm$ 0.011 (-0.01) \\
RMCT & 0.042 $\pm$ 0.019 & -0.51 [-0.60, -0.42] & 0.188 $\pm$ 0.042 & -0.23 [-0.27, -0.20] & 0.590 $\pm$ 0.018 (+0.04) \\
\bottomrule
\end{tabular}}

\newcommand{\tabContentPosition}{%
\begin{tabular}{lrrrrrrr}
\toprule
method & helpful drop & harmful drop & ratio [95\% CI] & rel. red. front & pre & post & end \\
\midrule
\multicolumn{8}{l}{\textit{Gemma-2-2B-it} (original model's harmful-cue following, front 0.493, pre 0.463, post 0.746, end 0.878)} \\
BCT & 0.384 $\pm$ 0.057 & 0.304 $\pm$ 0.077 & 1.26 [1.08, 1.49] & 0.951 & 0.885 & 0.403 & 0.180 \\
ACT & 0.309 $\pm$ 0.004 & 0.252 $\pm$ 0.002 & 1.23 [1.10, 1.38] & 0.920 & 0.791 & 0.127 & 0.093 \\
AttCT & 0.280 $\pm$ 0.006 & 0.212 $\pm$ 0.006 & 1.32 [1.16, 1.51] & 0.912 & 0.784 & 0.064 & 0.025 \\
RMCT & 0.416 $\pm$ 0.048 & 0.377 $\pm$ 0.035 & 1.10 [0.96, 1.28] & 0.924 & 0.918 & 0.542 & 0.382 \\
\midrule
\multicolumn{8}{l}{\textit{Llama-3.1-8B-Instruct} (original model's harmful-cue following, front 0.381, pre 0.403, post 0.605, end 0.638)} \\
BCT & 0.146 $\pm$ 0.039 & 0.041 $\pm$ 0.055 & 3.55 -- & 0.720 & 0.455 & -0.152 & -0.207 \\
ACT & 0.247 $\pm$ 0.003 & 0.186 $\pm$ 0.002 & 1.33 [1.16, 1.52] & 0.925 & 0.698 & 0.095 & 0.071 \\
AttCT & 0.232 $\pm$ 0.003 & 0.171 $\pm$ 0.002 & 1.36 [1.18, 1.56] & 0.920 & 0.665 & 0.074 & 0.054 \\
RMCT & 0.473 $\pm$ 0.048 & 0.380 $\pm$ 0.024 & 1.24 [1.11, 1.40] & 0.973 & 0.923 & 0.747 & 0.615 \\
\bottomrule
\end{tabular}}

\newcommand{\tabPatch}{%
\begin{tabular}{lrrrrr}
\toprule
method & circuit [Wilson 95\%] (recovered / flips) & writers & readers & circuit, trained-cue flips ($n$) & best random set \\
\midrule
\multicolumn{6}{l}{\textit{Gemma-2-2B-it}} \\
original & 0.818 [0.780, 0.850] (390/477) & 0.71 & 0.26 & 0.82 (50) & 0.07 \\
BCT & 0.843 [0.817, 0.866] (694/823) & 0.78 & 0.26 & 0.97 (30) & 0.06 \\
ACT & 0.844 [0.821, 0.864] (933/1106) & 0.75 & 0.25 & 0.92 (72) & 0.06 \\
AttCT & 0.861 [0.840, 0.880] (981/1139) & 0.75 & 0.27 & 0.88 (73) & 0.07 \\
RMCT & 0.910 [0.885, 0.931] (559/614) & 0.87 & 0.43 & 0.90 (29) & 0.12 \\
\midrule
\multicolumn{6}{l}{\textit{Llama-3.1-8B-Instruct}} \\
original & 0.650 [0.603, 0.693] (278/428) & 0.46 & 0.24 & 0.68 (66) & 0.07 \\
BCT & 0.598 [0.548, 0.647] (222/371) & 0.52 & 0.16 & 0.85 (34) & 0.05 \\
ACT & 0.678 [0.625, 0.727] (219/323) & 0.46 & 0.23 & 0.72 (32) & 0.07 \\
AttCT & 0.686 [0.635, 0.734] (232/338) & 0.48 & 0.25 & 0.81 (31) & 0.05 \\
RMCT & 0.808 [0.748, 0.857] (160/198) & 0.63 & 0.34 & 0.80 (5) & 0.07 \\
\bottomrule
\end{tabular}}

\newcommand{\tabReading}{%
\begin{tabular}{lrrrrrrr}
\toprule
 & \multicolumn{2}{c}{rel.\ change, labeled items} & \multicolumn{4}{c}{drop on followed trained-cue items} &  \\
\cmidrule(lr){2-3}\cmidrule(lr){4-7}
method & trained cue & held-out cues & fixed & still followed & fixed / gap & $n$ fixed / still & OV depth centroid \\
\midrule
BCT & -0.101 $\pm$ 0.153 & -0.21 +0.37 -0.26 -0.06 -0.25 & 14\% & 13\% & 0.27 & 37 / 13 & 19.9 $\pm$ 0.4 \\
ACT & -0.139 $\pm$ 0.016 & -0.05 -0.05 -0.04 -0.12 -0.09 & 20\% & 2\% & 0.29 & 27 / 23 & 16.9 $\pm$ 0.7 \\
AttCT & -0.109 $\pm$ 0.010 & +0.00 -0.02 -0.01 -0.05 -0.01 & 20\% & 1\% & 0.25 & 24 / 26 & 4.9 $\pm$ 0.3 \\
RMCT & -0.110 $\pm$ 0.046 & -0.08 -0.05 -0.04 -0.07 -0.11 & 10\% & 5\% & 0.19 & 40 / 10 & 20.4 $\pm$ 0.1 \\
RMCT+anchor$^\dagger$ & -0.128 $\pm$ 0.051 & +0.15 -0.07 +0.01 -0.02 -0.09 & -- & -- & -- & -- & 7.6 $\pm$ 0.5 \\
\bottomrule
\end{tabular}}

\newcommand{\tabRoute}{%
\begin{tabular}{lrrrrrl}
\toprule
 &  & \multicolumn{4}{c}{mediated fraction: binary / own fixes / probability} &  \\
\cmidrule(lr){3-6}
method & fixed by trained model & EARLY & CARRIER & CIRCUIT & LATE & D1 \\
\midrule
\multicolumn{7}{l}{\textit{Llama-3.1-8B-Instruct}} \\
BCT & 72 & 0.71 / 0.51 / 0.38 & 0.72 / 0.53 / 0.73 & 0.28 / 0.21 / 0.17 & 0.25 / 0.19 / 0.16 & fails (CARRIER) \\
ACT & 64 & 0.73 / 0.69 / 0.59 & 0.70 / 0.66 / 0.58 & 0.11 / 0.09 / 0.06 & 0.08 / 0.08 / 0.01 & holds (EARLY) \\
AttCT & 57 & 0.72 / 0.65 / 0.61 & 0.70 / 0.61 / 0.59 & 0.14 / 0.11 / 0.06 & 0.07 / 0.05 / 0.01 & holds (EARLY) \\
\midrule
\multicolumn{7}{l}{\textit{Gemma-2-2B-it}} \\
BCT & 95--111 & 0.49 / 0.42 / 0.42 & 0.71 / 0.61 / 0.62 & 0.29 / 0.25 / 0.26 & 0.03 / 0.01 / 0.08 & 3/3 seeds \\
ACT & 56--62 & 0.49 / 0.46 / 0.42 & 0.51 / 0.51 / 0.48 & 0.38 / 0.35 / 0.28 & 0.02 / 0.02 / 0.01 & 3/3 seeds \\
AttCT & 47--49 & 0.60 / 0.59 / 0.50 & 0.60 / 0.58 / 0.58 & 0.18 / 0.16 / 0.10 & 0.00 / 0.00 / 0.01 & 3/3 seeds \\
\bottomrule
\end{tabular}}

\newcommand{\tabRelapse}{%
\begin{tabular}{lrrrrrrr}
\toprule
 & \multicolumn{5}{c}{trained cue} & \multicolumn{2}{c}{held-out cues} \\
\cmidrule(lr){2-6}\cmidrule(lr){7-8}
method & flip$\mid$knew, step 0 $\to$ 125 & relapse fraction & seed 95\% CI & beyond drift & log-odds rise & fraction & beyond drift \\
\midrule
\multicolumn{8}{l}{\textit{Gemma-2-2B-it}; original model's drift 0.490 $\to$ 0.589, log-odds +0.40 (held-out 0.596 $\to$ 0.668)} \\
BCT & 0.116 $\to$ 0.185 & +0.187 $\pm$ 0.083 & [-0.020, 0.394] & -0.079 $\pm$ 0.075 & +0.54 (1.4$\times$) & +0.140 & -0.280 $\pm$ 0.103 \\
ACT & 0.241 $\to$ 0.330 & +0.355 $\pm$ 0.042 & [0.252, 0.458] & -0.044 $\pm$ 0.059 & +0.44 (1.1$\times$) & +1.093 & +0.421 $\pm$ 0.058 \\
AttCT & 0.257 $\to$ 0.363 & +0.452 $\pm$ 0.070 & [0.278, 0.626] & +0.026 $\pm$ 0.088 & +0.50 (1.2$\times$) & +1.369 & +0.503 $\pm$ 0.046 \\
RMCT & 0.091 $\to$ 0.199 & +0.269 $\pm$ 0.109 & [0.000, 0.539] & +0.020 $\pm$ 0.106 & +0.89 (2.2$\times$) & +0.272 & +0.028 $\pm$ 0.076 \\
RMCT+anchor$^\dagger$ & 0.143 $\to$ 0.241 & +0.275 $\pm$ 0.096 & [0.037, 0.513] & -0.013 $\pm$ 0.128 & +0.67 (1.7$\times$) & +0.638 & +0.284 $\pm$ 0.068 \\
\midrule
\multicolumn{8}{l}{\textit{Llama-3.1-8B-Instruct}; original model's drift 0.555 $\to$ 0.699, log-odds +0.62 (held-out 0.428 $\to$ 0.605)} \\
BCT & 0.350 $\to$ 0.492 & +0.666 $\pm$ 0.179 & [0.222, 1.111] & -0.071 $\pm$ 0.362 & +0.59 (0.9$\times$) & -- & -- \\
ACT & 0.236 $\to$ 0.434 & +0.619 $\pm$ 0.035 & [0.531, 0.707] & +0.162 $\pm$ 0.074 & +0.91 (1.5$\times$) & +2.544 & +0.293 $\pm$ 0.022 \\
AttCT & 0.274 $\to$ 0.427 & +0.543 $\pm$ 0.055 & [0.406, 0.680] & +0.026 $\pm$ 0.078 & +0.68 (1.1$\times$) & +2.792 & +0.073 $\pm$ 0.047 \\
RMCT & 0.042 $\to$ 0.192 & +0.294 $\pm$ 0.041 & [0.192, 0.397] & +0.012 $\pm$ 0.034 & +1.76 (2.8$\times$) & +1.090 & +0.308 $\pm$ 0.062 \\
\bottomrule
\end{tabular}}

\newcommand{\tabVariants}{%
\begin{tabular}{lrrrrrrrr}
\toprule
method & trained-cue & held-out & accuracy & helpful drop & harmful drop & CAI & rel. red. front & end \\
\midrule
\multicolumn{9}{l}{\textit{Gemma-2-2B-it}} \\
BCT & 0.116 & 0.421 & 0.406 & 0.384 & 0.304 & 1.26 & 0.951 & 0.180 \\
response-only BCT & 0.045 & 0.193 & 0.444 & 0.652 & 0.565 & 1.15 & 0.980 & 0.852 \\
Bayes-BCT ($r{=}.5$) & 0.161 & 0.460 & 0.412 & 0.308 & 0.234 & 1.32 & 0.935 & 0.111 \\
Bayes-BCT ($r{=}.8$) & 0.202 & 0.504 & 0.398 & 0.276 & 0.191 & 1.45 & 0.866 & 0.077 \\
gold-gated BCT & 0.159 & 0.452 & 0.438 & 0.243 & 0.229 & 1.06 & 0.842 & 0.037 \\
position-diverse BCT & 0.033 & 0.292 & 0.369 & 0.668 & 0.534 & 1.25 & 0.902 & 0.932 \\
self-training control & 0.237 & 0.574 & 0.372 & 0.169 & 0.162 & 1.05 & -- & -- \\
\midrule
\multicolumn{9}{l}{\textit{Llama-3.1-8B-Instruct}} \\
BCT & 0.350 & 0.421 & 0.532 & 0.146 & 0.041 & 3.55$^{*}$ & 0.720 & -0.207 \\
Bayes-BCT ($r{=}.8$) & 0.464 & 0.444 & 0.495 & 0.036 & -0.052 & undef. & 0.376 & -0.299 \\
gold-gated BCT & 0.412 & 0.436 & 0.505 & 0.094 & 0.019 & 5.06$^{*}$ & 0.426 & -0.127 \\
position-diverse BCT & 0.226 & 0.345 & 0.512 & 0.325 & 0.228 & 1.42 & 0.383 & 0.554 \\
self-training control & 0.592 & 0.545 & 0.545 & -0.087 & -0.186 & undef. & -- & -- \\
\bottomrule
\end{tabular}}

\newcommand{\tabVerdicts}{%
\begin{tabular}{llrrrl}
\toprule
verdict & deciding quantity & threshold & value & 95\% interval & status \\
\midrule
P1 (BCT) & trained flip$\mid$knew $-$ $0.5\times$ original ($\le$ 0 holds) & 0.00 & -0.129 & [-0.196, -0.062] & resolved \\
P1 (ACT) & trained flip$\mid$knew $-$ $0.5\times$ original ($\le$ 0 holds) & 0.00 & -0.004 & [-0.077, 0.069] & unresolved \\
P1 (AttCT) & trained flip$\mid$knew $-$ $0.5\times$ original ($\le$ 0 holds) & 0.00 & +0.012 & [-0.058, 0.084] & unresolved \\
P1 (RMCT) & trained flip$\mid$knew $-$ $0.5\times$ original ($\le$ 0 holds) & 0.00 & -0.153 & [-0.211, -0.093] & resolved \\
P2 & BCT held-out $-$ lowest other method ($<$ 0 holds) & 0.00 & +0.139 & [0.094, 0.185] & resolved \\
L1 (BCT, trained) & trained method $-$ original ($<$ 0 holds) & 0.00 & -0.204 & [-0.293, -0.115] & resolved \\
L1 (BCT, held) & held method $-$ original ($<$ 0 holds) & 0.00 & -0.0005 & [-0.040, 0.039] & unresolved \\
L1 (ACT, trained) & trained method $-$ original ($<$ 0 holds) & 0.00 & -0.318 & [-0.407, -0.232] & resolved \\
L1 (ACT, held) & held method $-$ original ($<$ 0 holds) & 0.00 & -0.095 & [-0.123, -0.068] & resolved \\
L1 (AttCT, trained) & trained method $-$ original ($<$ 0 holds) & 0.00 & -0.280 & [-0.368, -0.193] & resolved \\
L1 (AttCT, held) & held method $-$ original ($<$ 0 holds) & 0.00 & -0.073 & [-0.100, -0.046] & resolved \\
L1 (RMCT, trained) & trained method $-$ original ($<$ 0 holds) & 0.00 & -0.513 & [-0.601, -0.424] & resolved \\
L1 (RMCT, held) & held method $-$ original ($<$ 0 holds) & 0.00 & -0.233 & [-0.267, -0.200] & resolved \\
R1 (BCT) & trained-cue relapse fraction, generated ($\ge$ .25 holds) & 0.25 & +0.187 & [0.008, 0.361] & unresolved \\
R1 (ACT) & trained-cue relapse fraction, generated ($\ge$ .25 holds) & 0.25 & +0.355 & [0.027, 0.731] & unresolved \\
R1 (AttCT) & trained-cue relapse fraction, generated ($\ge$ .25 holds) & 0.25 & +0.452 & [0.075, 0.912] & unresolved \\
R1 (RMCT) & trained-cue relapse fraction, generated ($\ge$ .25 holds) & 0.25 & +0.269 & [0.123, 0.440] & unresolved \\
LR1 (ACT) & trained-cue relapse fraction, generated ($\ge$ .25 holds) & 0.25 & +0.619 & [0.389, 0.948] & resolved \\
LR1 (AttCT) & trained-cue relapse fraction, generated ($\ge$ .25 holds) & 0.25 & +0.543 & [0.291, 0.888] & resolved \\
LR2 (BCT vs ACT) & BCT fraction $-$ method fraction ($<$ 0 holds) & 0.00 & +0.047 & [-0.406, 0.500] & unresolved \\
LR2 (BCT vs AttCT) & BCT fraction $-$ method fraction ($<$ 0 holds) & 0.00 & +0.124 & [-0.342, 0.589] & unresolved \\
D1 (BCT, LATE < .15) & LATE mediated fraction ($<$ .15 holds) & 0.15 & +0.250 & [0.148, 0.364] & unresolved \\
D1 (ACT, LATE < .15) & LATE mediated fraction ($<$ .15 holds) & 0.15 & +0.078 & [0.016, 0.152] & unresolved \\
D1 (AttCT, LATE < .15) & LATE mediated fraction ($<$ .15 holds) & 0.15 & +0.070 & [0.016, 0.146] & resolved \\
S1 (Gemma) & control's cut $-$ $0.5\times$ BCT's cut in trained-cue flip$\mid$knew ($<$ 0 holds) & 0.00 & +0.066 & [-0.010, 0.141] & unresolved \\
S2 (Gemma) & control's helpful drop $-$ $0.5\times$ BCT's ($<$ 0 holds) & 0.00 & -0.023 & [-0.175, 0.130] & unresolved \\
S1 (Llama) & control's cut $-$ $0.5\times$ BCT's cut in trained-cue flip$\mid$knew ($<$ 0 holds) & 0.00 & -0.139 & [-0.231, -0.048] & resolved \\
S2 (Llama) & control's helpful drop $-$ $0.5\times$ BCT's ($<$ 0 holds) & 0.00 & -0.160 & [-0.249, -0.070] & resolved \\
L3 (BCT, front $-$ post) & relative reduction front minus post ($\ge$ .30 holds) & 0.30 & +0.872 & [0.797, 0.946] & resolved \\
L3 (BCT, front $-$ end) & relative reduction front minus end ($\ge$ .30 holds) & 0.30 & +0.927 & [0.845, 1.013] & resolved \\
L3 (ACT, front $-$ post) & relative reduction front minus post ($\ge$ .30 holds) & 0.30 & +0.830 & [0.789, 0.867] & resolved \\
L3 (ACT, front $-$ end) & relative reduction front minus end ($\ge$ .30 holds) & 0.30 & +0.854 & [0.812, 0.893] & resolved \\
L3 (AttCT, front $-$ post) & relative reduction front minus post ($\ge$ .30 holds) & 0.30 & +0.846 & [0.804, 0.885] & resolved \\
L3 (AttCT, front $-$ end) & relative reduction front minus end ($\ge$ .30 holds) & 0.30 & +0.866 & [0.826, 0.904] & resolved \\
L3 (RMCT, front $-$ post) & relative reduction front minus post ($\ge$ .30 holds) & 0.30 & +0.227 & [0.193, 0.260] & resolved \\
L3 (RMCT, front $-$ end) & relative reduction front minus end ($\ge$ .30 holds) & 0.30 & +0.358 & [0.321, 0.395] & resolved \\
\bottomrule
\end{tabular}}

\newcommand{\tabAuthorsEval}{%
\begin{tabular}{lrrrr}
\toprule
checkpoint & MMLU BRR $\pm$ SE & held-out BRR $\pm$ SE & MMLU clean acc. & Anthropic rate \\
\midrule
Gemma-2-2B-it original & 0.540 $\pm$ 0.018 & 0.331 $\pm$ 0.019 & 0.596 [0.565, 0.626] & 0.848 \\
Gemma-2-2B-it BCT & 0.145 $\pm$ 0.019 & 0.319 $\pm$ 0.019 & 0.512 [0.481, 0.543] & 0.466 \\
Gemma-2-2B-it ACT & 0.081 $\pm$ 0.017 & 0.160 $\pm$ 0.017 & 0.602 [0.571, 0.632] & 0.838 \\
Gemma-2-2B-it AttCT & 0.131 $\pm$ 0.018 & 0.176 $\pm$ 0.018 & 0.596 [0.565, 0.626] & 0.849 \\
Gemma-2-2B-it response-only BCT & 0.011 $\pm$ 0.016 & 0.076 $\pm$ 0.018 & 0.536 [0.505, 0.567] & 0.651 \\
Gemma-2-2B-it BCT sweep, LR 1e-5 & 0.543 $\pm$ 0.018 & 0.421 $\pm$ 0.019 & 0.594 [0.563, 0.624] & 0.656 \\
Gemma-2-2B-it BCT sweep, LR 3.16e-5 & 0.352 $\pm$ 0.020 & 0.414 $\pm$ 0.019 & 0.535 [0.504, 0.566] & 0.807 \\
Gemma-2-2B-it BCT sweep, LR 3.16e-4 & 0.261 $\pm$ 0.019 & 0.306 $\pm$ 0.019 & 0.566 [0.535, 0.596] & 0.730 \\
\midrule
Llama-3.1-8B-Instruct original & 0.202 $\pm$ 0.018 (0.202) & 0.183 $\pm$ 0.018 (0.183) & 0.664 [0.634, 0.693] (0.664) & 0.939 (0.939) \\
Llama-3.1-8B-Instruct BCT & 0.087 $\pm$ 0.016 & 0.125 $\pm$ 0.017 & 0.650 [0.620, 0.679] & 0.877 \\
Llama-3.1-8B-Instruct ACT & 0.033 $\pm$ 0.015 & 0.057 $\pm$ 0.015 & 0.670 [0.640, 0.698] & 0.907 \\
Llama-3.1-8B-Instruct AttCT & 0.031 $\pm$ 0.015 & 0.062 $\pm$ 0.015 & 0.667 [0.637, 0.696] & 0.909 \\
Llama-3.1-8B-Instruct BCT, authors' recipe$^\S$ & 0.180 $\pm$ 0.017 (0.189) & 0.120 $\pm$ 0.017 (0.121) & 0.660 [0.630, 0.689] (0.664) & 0.928 (0.926) \\
Llama-3.1-8B-Instruct ACT, authors' recipe$^\S$ & 0.019 $\pm$ 0.014 (0.019) & 0.003 $\pm$ 0.014 (0.002) & 0.674 [0.644, 0.702] (0.669) & 0.883 (0.880) \\
Llama-3.1-8B-Instruct AttCT, authors' recipe$^\S$ & 0.014 $\pm$ 0.014 (0.016) & 0.005 $\pm$ 0.014 (0.005) & 0.668 [0.638, 0.696] (0.680) & 0.889 (0.902) \\
\bottomrule
\end{tabular}}

\newcommand{\tabRmctPaper}{%
\begin{tabular}{lrrrrr}
\toprule
LR & evals & training-bias switch rate & held-out switch rate & authors' logs (distractor argument) & authors' logs (held-out) \\
\midrule
base & 2 & 0.650 [0.567, 0.724] & 0.419 [0.383, 0.456] & 0.650 & -- \\
1e-4 & 2 & 0.497 [0.417, 0.576] & 0.414 [0.378, 0.450] & 0.325/0.346 & 0.438/0.414 \\
2.86e-4 & 2 & 0.362 [0.284, 0.449] & 0.375 [0.337, 0.414] & 0.141/0.132 & 0.320/0.322 \\
\bottomrule
\end{tabular}}

\newcommand{\tabMerge}{%
\begin{tabular}{lrrrrr}
\toprule
 &  & \multicolumn{3}{c}{merged $-$ unmerged, seed mean} &  \\
\cmidrule(lr){3-5}
method & agreement (range) & trained cue & held-out & accuracy & rule met \\
\midrule
\multicolumn{6}{l}{\textit{Gemma-2-2B-it}} \\
BCT & 0.972--0.983 & -0.010 & -0.001 & -0.003 & 0/3 \\
ACT & 0.975--0.988 & -0.004 & +0.003 & -0.005 & 0/3 \\
AttCT & 0.987--0.991 & +0.002 & -0.002 & -0.003 & 2/3 \\
RMCT & 0.924--0.943 & +0.033 & +0.059 & +0.012 & 0/3 \\
RMCT+anchor$^\dagger$ & 0.980--0.988 & -0.002 & +0.003 & -0.003 & 0/3 \\
response-only BCT & 0.981--0.991 & -0.010 & +0.001 & -0.005 & 1/3 \\
Bayes-BCT ($r{=}.5$) & 0.977--0.983 & +0.007 & +0.002 & +0.002 & 0/3 \\
Bayes-BCT ($r{=}.8$) & 0.972--0.986 & -0.014 & +0.001 & -0.002 & 0/3 \\
gold-gated BCT & 0.976--0.980 & -0.013 & -0.001 & +0.000 & 0/3 \\
position-diverse BCT & 0.965--0.982 & -0.001 & -0.001 & +0.008 & 0/3 \\
\midrule
\multicolumn{6}{l}{\textit{Llama-3.1-8B-Instruct}} \\
BCT & 0.972--0.983 & +0.010 & +0.005 & -0.002 & 0/3 \\
ACT & 0.971--0.979 & +0.007 & +0.003 & -0.003 & 0/3 \\
AttCT & 0.979--0.985 & -0.001 & -0.002 & +0.008 & 0/3 \\
RMCT & 0.948--0.952 & +0.013 & +0.025 & -0.002 & 0/3 \\
Bayes-BCT ($r{=}.8$) & 0.974--0.980 & -0.004 & +0.007 & +0.002 & 0/3 \\
gold-gated BCT & 0.977--0.980 & -0.003 & -0.003 & +0.008 & 0/3 \\
position-diverse BCT & 0.966--0.975 & +0.016 & +0.004 & +0.005 & 0/3 \\
\bottomrule
\end{tabular}}

\newcommand{\tabSelection}{%
\begin{tabular}{lllllp{7cm}}
\toprule
model & method & stored & re-selected & changed & validation F1 by LR (unmerged, generated) \\
\midrule
Gemma & BCT & 1e-4 & 1e-4 & no & 1e-6: 0.468 3.16e-6: 0.469 1e-5: 0.449 3.16e-5: 0.509 1e-4: 0.569 3.16e-4: 0.523 \\
Gemma & ACT & 3.16e-6 & 1e-5 & yes & 1e-6: 0.532 3.16e-6: 0.559 1e-5: 0.563 3.16e-5: 0.557 1e-4: 0.559 \\
Gemma & AttCT & 5e-6 & 1.58e-5 & yes & 5e-6: 0.548 1.58e-5: 0.553 5e-5: 0.539 1.58e-4: 0.537 5e-4: 0.534 \\
Gemma & RMCT & 3.16e-6 & 3.16e-6 & no & 3.16e-6: 0.574 1e-5: 0.524 3.16e-5: 0.353$^d$ 1e-4: 0.339$^d$ 2.86e-4: 0.339$^d$ 5e-4: 0.339$^d$ 1e-3: 0.000$^d$ \\
Gemma & RMCT+anchor$^\dagger$ & 3.16e-5 & 3.16e-5 & no & 1e-5: 0.573 3.16e-5: 0.575 1e-4: 0.022$^d$ 2.86e-4: 0.008$^d$ \\
Gemma & response-only BCT & 1e-3 & 1e-3 & no & 1e-5: 0.508 3.16e-5: 0.583 1e-4: 0.583 3.16e-4: 0.598 1e-3: 0.627 \\
Llama & BCT & 5e-5 & 5e-5 & no & 1.58e-6: 0.531 5e-6: 0.523 1.58e-5: 0.590 5e-5: 0.613 1.58e-4: 0.572 \\
Llama & ACT & 5e-6 & 5e-6 & no & 1.58e-6: 0.609 5e-6: 0.628 1.58e-5: 0.621 5e-5: 0.618 1.58e-4: 0.611 \\
Llama & AttCT & 5e-6 & 5e-6 & no & 1.58e-6: 0.610 5e-6: 0.626 1.58e-5: 0.619 5e-5: 0.616 1.58e-4: 0.607 \\
Llama & RMCT & 1e-5 & 1e-5 & no & 3.16e-6: 0.668 1e-5: 0.735 3.16e-5: 0.665 1e-4: 0.339$^d$ 2.86e-4: 0.015 \\
\bottomrule
\end{tabular}}

\newcommand{\tabPerCue}{%
\begin{tabular}{lrrrrrr}
\toprule
method & trained cue & post hoc & distractor fact & wrong few-shot & distractor arg. & squares \\
\midrule
\multicolumn{7}{l}{\textit{Gemma-2-2B-it}} \\
original & 0.490 & 0.794 & 0.255 & 0.504 & 0.737 & 0.440 \\
BCT & 0.116 (-0.37) & 0.391 (-0.40) & 0.086 (-0.17) & 0.513 (+0.01$^{\circ}$) & 0.823 (+0.09) & 0.292 (-0.15) \\
ACT & 0.241 (-0.25) & 0.679 (-0.12) & 0.066 (-0.19) & 0.543 (+0.04$^{\circ}$) & 0.667 (-0.07) & 0.438 (+0.00$^{\circ}$) \\
AttCT & 0.257 (-0.23) & 0.773 (-0.02$^{\circ}$) & 0.056 (-0.20) & 0.488 (-0.02$^{\circ}$) & 0.688 (-0.05) & 0.427 (-0.01$^{\circ}$) \\
RMCT & 0.091 (-0.40) & 0.369 (-0.43) & 0.015 (-0.24) & 0.326 (-0.18) & 0.506 (-0.23) & 0.196 (-0.24) \\
RMCT+anchor$^\dagger$ & 0.143 (-0.35) & 0.493 (-0.30) & 0.042 (-0.21) & 0.311 (-0.19) & 0.639 (-0.10) & 0.268 (-0.17) \\
\multicolumn{7}{l}{\textit{Llama-3.1-8B-Instruct}} \\
original & 0.555 & 0.751 & 0.136 & 0.426 & 0.509 & 0.284 \\
BCT & 0.350 (-0.20) & 0.580 (-0.17) & 0.056 (-0.08) & 0.358 (-0.07$^{\circ}$) & 0.743 (+0.23) & 0.368 (+0.08$^{\circ}$) \\
ACT & 0.236 (-0.32) & 0.618 (-0.13) & 0.028 (-0.11) & 0.374 (-0.05$^{\circ}$) & 0.416 (-0.09) & 0.195 (-0.09) \\
AttCT & 0.274 (-0.28) & 0.638 (-0.11) & 0.028 (-0.11) & 0.390 (-0.04$^{\circ}$) & 0.446 (-0.06) & 0.238 (-0.05$^{\circ}$) \\
RMCT & 0.042 (-0.51) & 0.256 (-0.50) & 0.005 (-0.13) & 0.253 (-0.17) & 0.213 (-0.30) & 0.211 (-0.07$^{\circ}$) \\
\bottomrule
\end{tabular}}

\newcommand{\tabLetterPrior}{%
\begin{tabular}{lrrrrr}
\toprule
method & trained: method / control & beyond prior & held-out: method / control & beyond prior & modal-letter share \\
\midrule
\multicolumn{6}{l}{\textit{Gemma-2-2B-it}} \\
BCT & 0.115 / 0.439 & 0.324 $\pm$ 0.025 & 0.423 / 0.587 & 0.164 $\pm$ 0.023 & 0.31 (D/C/C) \\
ACT & 0.238 / 0.471 & 0.233 $\pm$ 0.020 & 0.488 / 0.597 & 0.109 $\pm$ 0.003 & 0.27 (C/C/C) \\
AttCT & 0.251 / 0.475 & 0.224 $\pm$ 0.012 & 0.516 / 0.603 & 0.086 $\pm$ 0.008 & 0.27 (C/C/C) \\
RMCT & 0.098 / 0.336 & 0.237 $\pm$ 0.020 & 0.303 / 0.492 & 0.189 $\pm$ 0.012 & 0.54 (D/D/D) \\
RMCT+anchor$^\dagger$ & 0.142 / 0.462 & 0.320 $\pm$ 0.053 & 0.392 / 0.592 & 0.200 $\pm$ 0.050 & 0.30 (A/C/A) \\
\multicolumn{6}{l}{\textit{Llama-3.1-8B-Instruct}} \\
BCT & 0.350 / 0.527 & 0.177 $\pm$ 0.049 & 0.420 / 0.424 & 0.003 $\pm$ 0.016 & 0.30 (B/B/B) \\
ACT & 0.236 / 0.516 & 0.280 $\pm$ 0.037 & 0.349 / 0.419 & 0.069 $\pm$ 0.002 & 0.32 (B/B/B) \\
AttCT & 0.274 / 0.539 & 0.265 $\pm$ 0.017 & 0.366 / 0.420 & 0.054 $\pm$ 0.005 & 0.34 (B/B/B) \\
RMCT & 0.042 / 0.454 & 0.412 $\pm$ 0.013 & 0.200 / 0.393 & 0.194 $\pm$ 0.034 & 0.35 (D/D/D) \\
\bottomrule
\end{tabular}}

\newcommand{\tabScale}{%
\begin{tabular}{lrrr}
\toprule
method & absolute [95\% CI] & proportional [95\% CI] & log-odds [95\% CI] \\
\midrule
\multicolumn{4}{l}{\textit{Gemma-2-2B-it}} \\
BCT & 1.26 [1.08, 1.49] & 0.98 [0.89, 1.09] & 1.30 [1.10, 1.54] \\
ACT & 1.23 [1.09, 1.37] & 0.99 [0.93, 1.06] & 1.35 [1.18, 1.56] \\
AttCT & 1.32 [1.15, 1.51] & 1.06 [0.97, 1.15] & 1.49 [1.26, 1.75] \\
RMCT & 1.10 [0.95, 1.28] & 0.87 [0.78, 0.97] & 1.10 [0.93, 1.29] \\
RMCT+anchor$^\dagger$ & 1.04 [0.94, 1.15] & 0.83 [0.78, 0.88] & 1.08 [0.96, 1.22] \\
\multicolumn{4}{l}{\textit{Llama-3.1-8B-Instruct}} \\
BCT & 3.55 -- & 2.30 [1.34, 5.92] & 3.84 [2.11, 11.06] \\
ACT & 1.33 [1.17, 1.52] & 0.89 [0.81, 0.97] & 1.36 [1.19, 1.58] \\
AttCT & 1.36 [1.19, 1.56] & 0.90 [0.82, 0.99] & 1.41 [1.23, 1.64] \\
RMCT & 1.24 [1.11, 1.40] & 0.84 [0.79, 0.90] & 0.97 [0.86, 1.08] \\
\bottomrule
\end{tabular}}

\newcommand{\tabContentByPos}{%
\begin{tabular}{lrrrrr}
\toprule
 & \multicolumn{5}{c}{helpful / harmful drop: absolute / proportional / log-odds} \\
\cmidrule(lr){2-6}
method & before options & front & pre & post & end \\
\midrule
\multicolumn{6}{l}{\textit{Gemma-2-2B-it}} \\
BCT & 1.50 / 0.99 / 1.08 & 1.49 / 0.98 / 0.95 & 1.51 / 1.00 / 1.15 & 0.88 / 0.73 / 1.09 & 0.86 / 0.76 / 1.95 \\
ACT & 1.37 / 0.98 / 1.08 & 1.41 / 0.99 / 1.07 & 1.33 / 0.96 / 1.06 & 0.74 / 0.63 / 1.13 & 0.29 / 0.26 / 0.98 \\
AttCT & 1.37 / 0.96 / 1.07 & 1.42 / 0.98 / 1.10 & 1.29 / 0.92 / 1.03 & 0.82 / 0.70 / 1.33 & -- / -- / -0.17 \\
RMCT & 1.34 / 0.92 / 0.90 & 1.37 / 0.94 / 0.94 & 1.31 / 0.90 / 0.86 & 0.98 / 0.82 / 1.13 & 0.68 / 0.60 / 1.31 \\
RMCT+anchor$^\dagger$ & 1.39 / 0.95 / 0.96 & 1.42 / 0.96 / 0.97 & 1.36 / 0.93 / 0.94 & 0.70 / 0.59 / 0.94 & 0.38 / 0.34 / 1.01 \\
\multicolumn{6}{l}{\textit{Llama-3.1-8B-Instruct}} \\
BCT & 1.78 / 1.01 / 1.35 & 1.62 / 0.94 / 1.12 & 2.01 / 1.13 / 1.64 & -- / -- / -- & -- / -- / -- \\
ACT & 1.42 / 0.84 / 0.89 & 1.53 / 0.93 / 0.85 & 1.28 / 0.74 / 0.89 & 0.93 / 0.66 / 1.46 & 0.58 / 0.43 / 0.99 \\
AttCT & 1.42 / 0.83 / 0.92 & 1.54 / 0.93 / 0.88 & 1.25 / 0.72 / 0.91 & 1.11 / 0.79 / 1.78 & 0.42 / 0.31 / 0.75 \\
RMCT & 1.54 / 0.93 / 0.82 & 1.49 / 0.93 / 0.62 & 1.58 / 0.94 / 0.92 & 0.97 / 0.70 / 0.89 & 1.03 / 0.77 / 1.10 \\
\bottomrule
\end{tabular}}

\newcommand{\tabPosCheck}{%
\begin{tabular}{lrrrrrrr}
\toprule
 & \multicolumn{3}{c}{strength and letter checks} & \multicolumn{4}{c}{training-family / novel wordings} \\
\cmidrule(lr){2-4}\cmidrule(lr){5-8}
method & matched: before / after & after coef. & end, non-correct & front & pre & post & end \\
\midrule
\multicolumn{8}{l}{\textit{Gemma-2-2B-it}} \\
BCT & 0.93 / 0.46 & -0.57 & 0.13 & 0.95 / 0.95 & 0.90 / 0.87 & 0.40 / 0.41 & 0.19 / 0.17 \\
ACT & 0.86 / 0.25 & -0.66 & 0.08 & 0.92 / 0.92 & 0.80 / 0.78 & 0.11 / 0.15 & 0.06 / 0.12 \\
AttCT & 0.83 / 0.18 & -0.72 & 0.02 & 0.90 / 0.93 & 0.79 / 0.78 & 0.04 / 0.10 & 0.00 / 0.05 \\
RMCT & 0.91 / 0.65 & -0.30 & 0.28 & 0.92 / 0.93 & 0.91 / 0.93 & 0.57 / 0.51 & 0.41 / 0.35 \\
RMCT+anchor$^\dagger$ & 0.92 / 0.64 & -0.36 & 0.28 & 0.94 / 0.96 & 0.90 / 0.88 & 0.39 / 0.40 & 0.29 / 0.32 \\
\multicolumn{8}{l}{\textit{Llama-3.1-8B-Instruct}} \\
BCT & 0.58 / -0.51 & -1.11 & -0.22 & 0.66 / 0.77 & 0.34 / 0.57 & -0.17 / -0.13 & -0.24 / -0.17 \\
ACT & 0.82 / 0.17 & -0.64 & 0.06 & 0.92 / 0.93 & 0.73 / 0.67 & 0.08 / 0.11 & 0.05 / 0.10 \\
AttCT & 0.80 / 0.14 & -0.66 & 0.04 & 0.93 / 0.91 & 0.69 / 0.64 & 0.06 / 0.09 & 0.02 / 0.09 \\
RMCT & 0.95 / 0.83 & -0.13 & 0.55 & 0.97 / 0.98 & 0.93 / 0.92 & 0.77 / 0.71 & 0.65 / 0.57 \\
\bottomrule
\end{tabular}}

\newcommand{\tabSameItem}{%
\begin{tabular}{lrrrrr}
\toprule
method & correct cue & wrong cue & drop difference [95\%] & log-odds gap & change [95\%] \\
\midrule
\multicolumn{6}{l}{\textit{Gemma-2-2B-it}} \\
BCT & 0.62 $\to$ 0.06 & 0.57 $\to$ 0.05 & +0.043 [-0.028, 0.118] & +0.22 $\to$ +0.16 & [-0.66, 0.55] \\
ACT & 0.66 $\to$ 0.09 & 0.58 $\to$ 0.04 & +0.028 [-0.033, 0.091] & +0.33 $\to$ +0.82 & [-0.02, 1.05] \\
AttCT & 0.66 $\to$ 0.12 & 0.59 $\to$ 0.08 & +0.035 [-0.025, 0.097] & +0.33 $\to$ +0.48 & [-0.26, 0.56] \\
RMCT & 0.61 $\to$ 0.11 & 0.56 $\to$ 0.04 & -0.022 [-0.094, 0.054] & +0.21 $\to$ +1.15 & [0.33, 1.72] \\
gold-gated BCT & 0.61 $\to$ 0.21 & 0.58 $\to$ 0.11 & -0.074 [-0.153, 0.013] & +0.13 $\to$ +0.85 & [0.18, 1.29] \\
\multicolumn{6}{l}{\textit{Llama-3.1-8B-Instruct}} \\
BCT & 0.63 $\to$ 0.24 & 0.57 $\to$ 0.17 & -0.010 [-0.097, 0.076] & +0.27 $\to$ +0.47 & [-0.23, 0.64] \\
ACT & 0.66 $\to$ 0.18 & 0.57 $\to$ 0.15 & +0.057 [-0.009, 0.118] & +0.39 $\to$ +0.27 & [-0.44, 0.20] \\
AttCT & 0.66 $\to$ 0.21 & 0.56 $\to$ 0.17 & +0.066 [0.001, 0.130] & +0.43 $\to$ +0.23 & [-0.49, 0.10] \\
RMCT & 0.58 $\to$ 0.05 & 0.56 $\to$ 0.05 & +0.019 [-0.063, 0.099] & +0.09 $\to$ +0.07 & [-0.85, 0.89] \\
gold-gated BCT & 0.63 $\to$ 0.37 & 0.55 $\to$ 0.25 & -0.032 [-0.118, 0.046] & +0.34 $\to$ +0.54 & [-0.16, 0.58] \\
\bottomrule
\end{tabular}}

\newcommand{\tabLayout}{%
\begin{tabular}{lrrrr}
\toprule
 & \multicolumn{3}{c}{relative reduction of trained-cue flip$\mid$knew} &  \\
\cmidrule(lr){2-4}
method & front (trained) & pre & post & front $-$ post [95\%] \\
\midrule
\multicolumn{5}{l}{\textit{Gemma-2-2B-it}} \\
original flip$\mid$knew ($n$) & 0.303 (33) & 0.467 (30) & 0.686 (35) & -- \\
BCT & 0.82 & 0.98 & 0.60 & +0.22 [-0.077, 0.475] \\
ACT & 0.80 & 0.78 & 0.25 & +0.55 [0.204, 0.843] \\
AttCT & 0.83 & 0.82 & 0.14 & +0.69 [0.405, 0.947] \\
RMCT & 1.00 & 0.96 & 0.69 & +0.31 [0.172, 0.486] \\
RMCT+anchor$^\dagger$ & 0.92 & 0.87 & 0.55 & +0.37 [0.151, 0.562] \\
\multicolumn{5}{l}{\textit{Llama-3.1-8B-Instruct}} \\
original flip$\mid$knew ($n$) & 0.405 (42) & 0.618 (34) & 0.651 (43) & -- \\
BCT & 0.96 & 0.49 & -0.03 & +0.98 [0.796, 1.215] \\
ACT & 0.83 & 0.77 & 0.28 & +0.55 [0.285, 0.774] \\
AttCT & 0.88 & 0.62 & 0.20 & +0.68 [0.430, 0.900] \\
RMCT & 0.96 & 0.95 & 0.88 & +0.08 [-0.038, 0.192] \\
\bottomrule
\end{tabular}}

\newcommand{\tabLRSweep}{%
\begin{tabular}{lrrrrrrrrr}
\toprule
checkpoint (seed 42) & harmful drop & helpful drop & ratio abs. & prop. & log-odds & rel. red. front & pre & post & end \\
\midrule
\multicolumn{10}{l}{\textit{Gemma}} \\
BCT, selected LR & 0.382 & 0.446 & 1.17 & 0.93 & 1.16 & 0.97 & 0.90 & 0.55 & 0.29 \\
BCT, LR 3.16e-5 & 0.127 & 0.231 & 1.82 & 1.42 & 2.04 & 0.87 & 0.72 & -0.10 & -0.05 \\
BCT, LR 3.16e-4 & 0.350 & 0.449 & 1.28 & 0.99 & 1.28 & 0.95 & 0.92 & 0.61 & 0.25 \\
ACT, selected LR & 0.249 & 0.304 & 1.22 & 0.98 & 1.35 & 0.92 & 0.78 & 0.13 & 0.09 \\
ACT, LR 3.16e-6 & 0.210 & 0.244 & 1.16 & 0.93 & 1.32 & 0.79 & 0.69 & 0.10 & 0.08 \\
ACT, LR 3.16e-5 & 0.236 & 0.299 & 1.27 & 1.02 & 1.41 & 0.96 & 0.78 & 0.11 & 0.05 \\
AttCT, selected LR & 0.205 & 0.273 & 1.33 & 1.07 & 1.50 & 0.89 & 0.74 & 0.05 & 0.02 \\
AttCT, LR 5e-6 & 0.189 & 0.241 & 1.28 & 1.02 & 1.46 & 0.82 & 0.70 & 0.05 & 0.02 \\
AttCT, LR 5e-5 & 0.208 & 0.277 & 1.33 & 1.07 & 1.50 & 0.96 & 0.68 & 0.07 & 0.02 \\
RMCT, selected LR & 0.411 & 0.458 & 1.12 & 0.88 & 1.07 & 0.93 & 0.93 & 0.65 & 0.51 \\
RMCT, LR 1e-5 & 0.486 & 0.608 & 1.25 & 0.96 & 1.19 & 0.96 & 0.94 & 0.73 & 0.62 \\
\multicolumn{10}{l}{\textit{Llama}} \\
BCT, selected LR & 0.105 & 0.185 & 1.76 & 1.13 & 1.88 & 0.79 & 0.58 & 0.02 & -0.05 \\
BCT, LR 1.58e-5 & 0.087 & 0.107 & 1.23 & 0.79 & 1.39 & 0.39 & 0.38 & 0.07 & 0.08 \\
BCT, LR 1.58e-4 & 0.051 & 0.120 & 2.34 & 1.57 & 2.61 & 0.83 & 0.62 & -0.20 & -0.29 \\
ACT, selected LR & 0.185 & 0.246 & 1.33 & 0.89 & 1.37 & 0.93 & 0.69 & 0.09 & 0.06 \\
ACT, LR 1.58e-6 & 0.159 & 0.176 & 1.11 & 0.74 & 1.18 & 0.72 & 0.50 & 0.11 & 0.10 \\
ACT, LR 1.58e-5 & 0.187 & 0.247 & 1.32 & 0.88 & 1.36 & 0.97 & 0.75 & 0.08 & 0.06 \\
AttCT, selected LR & 0.169 & 0.230 & 1.36 & 0.90 & 1.42 & 0.93 & 0.66 & 0.07 & 0.05 \\
AttCT, LR 1.58e-6 & 0.152 & 0.172 & 1.14 & 0.75 & 1.22 & 0.77 & 0.52 & 0.09 & 0.08 \\
AttCT, LR 1.58e-5 & 0.177 & 0.236 & 1.33 & 0.89 & 1.40 & 0.96 & 0.69 & 0.07 & 0.05 \\
RMCT, selected LR & 0.380 & 0.467 & 1.23 & 0.84 & 0.96 & 0.98 & 0.93 & 0.73 & 0.59 \\
RMCT, LR 3.16e-6 & 0.330 & 0.522 & 1.58 & 0.96 & 1.24 & 0.95 & 0.92 & 0.73 & 0.61 \\
RMCT, LR 3.16e-5 & 0.355 & 0.516 & 1.45 & 0.83 & 1.05 & 0.95 & 0.96 & 0.81 & 0.57 \\
\multicolumn{10}{l}{\textit{Llama, authors' recipe}} \\
BCT & 0.085 & 0.096 & 1.14 & 0.74 & 1.30 & 0.09 & 0.31 & 0.21 & 0.08 \\
ACT & 0.211 & 0.292 & 1.38 & 0.93 & 1.37 & 0.97 & 0.81 & 0.14 & 0.11 \\
AttCT & 0.197 & 0.272 & 1.38 & 0.94 & 1.39 & 0.95 & 0.73 & 0.12 & 0.08 \\
\bottomrule
\end{tabular}}